\documentclass[fleqn,10pt]{wlscirep}
\usepackage[utf8]{inputenc}
\usepackage{mdframed}
\pdfoutput=1
\usepackage[frozencache,cachedir=.]{minted}
\usepackage{longtable}
\usepackage{array}
\usepackage{amssymb}
\usepackage{subcaption}
\newcolumntype{L}{>{\centering\arraybackslash}m{3cm}}
\usepackage[T1]{fontenc}
\usepackage[htt]{hyphenat}

\usepackage[user,titleref]{zref}

\title{Creation and Validation of a Chest X-Ray Dataset with Eye-tracking and Report Dictation for AI Development}

\author[1, *]{Alexandros Karargyris}
\author[1,$\dag$, *]{Satyananda Kashyap}
\author[1,$\dag$]{Ismini Lourentzou}
\author[1,$\dag$]{Joy Wu}
\author[1]{Arjun Sharma}
\author[1]{Matthew Tong}
\author[1]{Shafiq Abedin}
\author[1]{David Beymer}
\author[1]{Vandana Mukherjee}
\author[2]{Elizabeth A Krupinski}
\author[1, *]{Mehdi Moradi}

\affil[1]{IBM Research, Almaden Research Center, San Jose, CA, 95120, USA}
\affil[2]{Department of Radiology and Imaging Sciences, Emory University, Atlanta, GA, 30322, USA}

\affil[*]{corresponding author(s): Alexandros Karargyris (akarargyris@gmail.com), Satyananda Kashyap (satyananda.kashyap@ibm.com), Mehdi Moradi (mmoradi@us.ibm.com)}

\affil[$\dag$]{these authors contributed equally to this work}

\usepackage{color}

\begin{abstract}
We developed a rich dataset of Chest X-Ray (CXR) images to assist investigators in artificial intelligence. The data were collected using an eye tracking system while a radiologist reviewed and reported on 1,083 CXR images. The dataset contains the following aligned data: CXR image, transcribed radiology report text, radiologist's dictation audio and eye gaze coordinates data. We hope this dataset can contribute to various areas of research particularly towards explainable and multimodal deep learning / machine learning methods. Furthermore, investigators in disease classification and localization, automated radiology report generation, and human-machine interaction can benefit from these data. We report deep learning experiments that utilize the attention maps produced by eye gaze dataset to show the potential utility of this data. 
\end{abstract}
\begin{document}

\flushbottom
\maketitle

\thispagestyle{empty}

\section*{Background \& Summary}

In recent years, artificial intelligence (AI) has been extensively explored for enhancing the efficacy and efficiency of the radiology interpretation and reporting process. As the current prevalent paradigm of AI is deep learning, many of the works in AI for radiology use large data sets of labelled radiology images to train deep neural networks to classify images according to disease classes. Given the high labor cost of annotating images with the areas depicting the disease, large public training datasets often come with global labels describing the whole image \cite{irvin2019chexpert, johnson2019mimic} without localized annotation of the disease areas. The deep neural network model is trusted with discovering the relevant part of the image and learning the features characterizing the disease. This limits the performance of the resulting network. Furthermore, the black-box nature of deep neural networks and lack of local annotations means that the process of developing disease classifiers does not take advantage of experts' knowledge of disease appearance and location in medical images. The result is a multi-layer and nonlinear model with serious concerns with respect to explainability of its output. Furthermore, the generalization capability, (i.e., when the model is deployed to infer the class labels for images from other sources or distributions) is affected by scanner differences and/or demographic changes is a well-studied concern \cite{bluemke2020assessing}.

In the past five decades eye tracking has been extensively used in radiology for education, perception understanding, and fatigue measurement (example reviews: \cite{waite2019analysis,van2017visual,krupinski2010current,tourassi2013investigating}). More recently, efforts  \cite{khosravan2019collaborative, stember2019eye, aresta2020automatic, mall2018modeling} have used eye tracking data to improve segmentation and disease classification in Computed Tomography (CT) radiography by integrating them in deep learning techniques.
With such evidence and with the lack of public datasets that capture eye gaze data in the chest X-Ray (CXR) space, we present a new dataset that can help improve the way machine learning models are developed for radiology applications and we demonstrate its use in some popular deep learning architectures. 

This dataset consists of eye gaze information recorded from a single radiologist interpreting frontal chest radiographs. Dictation data (audio and timestamped text) of the radiology report reading is also provided. We also generated bounding boxes containing anatomical structures on every image and share them as part of this dataset. These bounding boxes can be used in conjunction with eye gaze information to produce more meaningful analyses. 

\noindent We present evidence that this dataset can help with two important tasks for AI practitioners in radiology:

\begin{itemize}
\item
The coordinates marking the areas of the image that a radiologist looks at while reporting a finding provide an approximate region of interest/attention for that finding. Without altering a radiologist's routine, this approach presents an inexpensive and efficient method for generating a locally annotated collection of images for training machine learning algorithms (e.g., disease classifiers). Since we also share the ground truth bounding boxes, the validity of the eye tracking in marking the location of the finding can be further studied using this dataset. We demonstrate utilization of eye gaze in deep neural network training and show an improvement in performance can be obtained. 
\item
Tracking of the eyes can characterize how radiologists approach the task of reading radiographs. Study of the eye-gaze of radiologists while reading normal and disease radiographs, presented as attention maps, reveals a cognitive workflow pattern that AI developers can use when building their models. 
\end{itemize}

We invite researchers in the radiology community who wish to contribute to the further development of the dataset to contact us.

\section*{Methods}\zlabel{Methods}

\begin{figure}[!ht]
\centering
\includegraphics[scale=0.5]{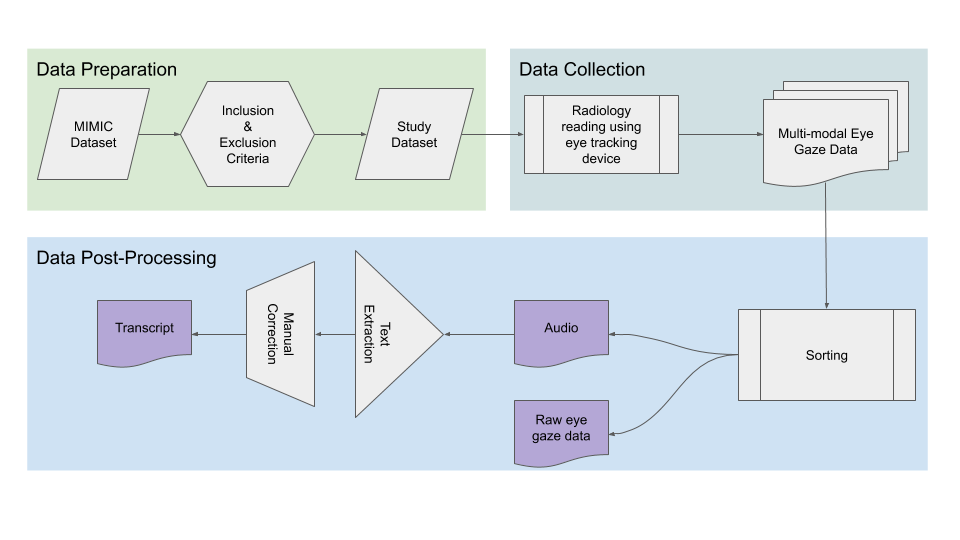}
\caption{Flowchart of Study}
\label{fig:study-design}
\end{figure}

Figure \ref{fig:study-design} provides an overview of the study and data generation process. For this study we used the publicly available MIMIC-CXR Database \cite{johnson2019mimic, goldberger2000physiobank} in conjunction with the publicly available Emergency Department (ED) subset of the MIMIC-IV Clinical Database \cite{johnson2020mimic4}. The MIMIC-IV-ED subset contains clinical observations/data and outcomes related to some of the CXR exams in the MIMIC-CXR database. Inclusion and exclusion criteria were applied to the patient attributes and clinical outcomes (via the discharge diagnosis, a.k.a the ICD-9 code) recorded in the MIMIC-IV Clinical Database,\cite{johnson2020mimic4} resulting in a subset of 1,083 cases that equally cover 3 conditions: Normal, Pneumonia and Congestive Heart Failure (CHF). The corresponding CXR images of these cases were extracted from the MIMIC-CXR database \cite{johnson2019mimic}. A radiologist (American Board of Radiology certified with over 5 years of experience) performed routine radiology reading of the images using the Gazepoint GP3 Eye Tracker\cite{GazepointHardware} (i.e., eye tracking device), Gazepoint Analysis UX Edition software\cite{GazepointSoftware} (i.e., software for performing eye gaze experiments), a headset microphone, a desktop computer and a monitor (Dell S2719DGF) set at 1920x1080 resolution. Radiology reading took place in multiple sessions (i.e., 30 cases per session) over a period of 2 months (i.e., March - May 2020). The Gazepoint Analysis UX Edition\cite{GazepointSoftware} exported video files (.avi format) containing eye fixations and voice dictation of radiologist's reading along with spreadsheets (.csv format) containing eye tracker's recorded eye gaze data. The audio was extracted from the video files and saved in \texttt{wav} and  \texttt{mp3} format. Subsequently, these audio files were processed with speech-to-text software (i.e., \href{https://cloud.google.com/speech-to-text}{Google Speech-to-Text}) to extract text transcripts along with dictation word time-related information (.json format). Furthermore, these transcripts were manually corrected. The final dataset contains the raw eye gaze signal information (.csv), audio files (.wav, .mp3) and transcript files (.json).

\subsection*{Ethical Statement}
The source data from the MIMIC-CXR and MIMIC-IV databases have been previously de-identified, and the institutional review boards of the Massachusetts Institute of Technology (No. 0403000206) and Beth Israel Deaconess Medical Center (2001-P-001699/14) both approved the use of the databases for research. We have also complied with all relevant ethical regulations with the use of the data for our study.

\noindent 

\subsection*{Data Preparation}\zlabel{data_preparation}
\subsubsection*{Inclusion and Exclusion Criteria}
\begin{figure}[!ht]
\centering
\includegraphics[scale=0.7]{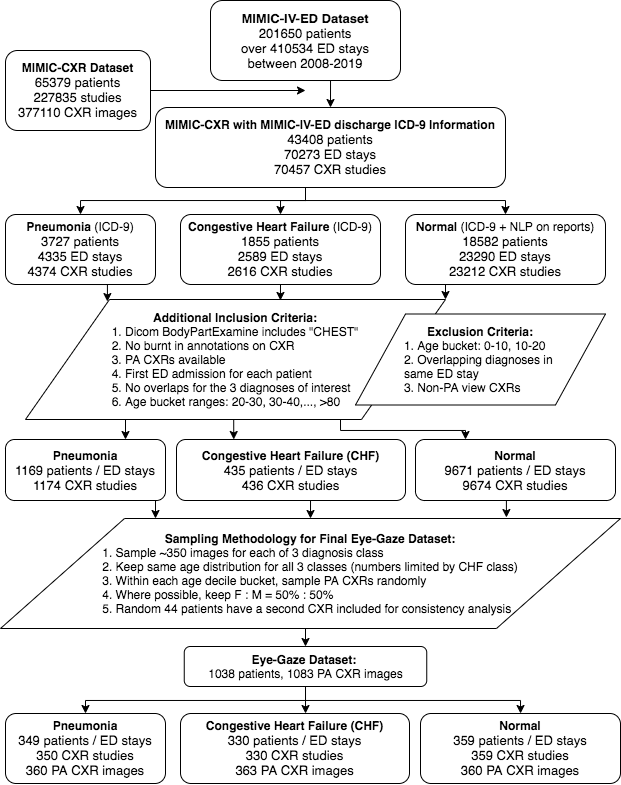}
\caption{Sampling flowchart for selecting images for this study from the MIMIC-IV (the ED subset) and the MIMIC-CXR datasets.\cite{johnson2020mimic4,johnson2019mimic}}
\label{fig:datasetflowchart}
\end{figure}

\zlabel{Inclusion/Exclusion}
Figure \ref{fig:datasetflowchart} describes the inclusion/exclusion criteria used to generate this dataset. These criteria were applied on the MIMIC-IV Clinical Database \cite{johnson2020mimic4} to identify the CXR studies of interest. The studies were used to extract their corresponding CXR images from the MIMIC-CXR Database \cite{johnson2019mimic}. 

We selected two clinically prevalent and high impact diseases, pneumonia and congestive heart failure (CHF), in the Emergency Department (ED) setting. We also picked normal cases as a comparison class. Unlike related CXR labeling efforts,\cite{irvin2019chexpert} where the same labels are derived from radiology reports using natural language processing (NLP) alone, the ground truth for our pneumonia and CHF class labels were derived from unique discharge ICD-9 codes (verified by our clinicians) from the MIMIC-IV-ED tables.\cite{johnson2020mimic4}

This ensures the ground truth is based on a formal clinical diagnosis and is likely to be more reliable, given that ICD-9 discharge diagnoses are typically derived from a multi-disciplinary team of treating providers after having considered all clinically relevant information (e.g., bedside observations, labs) in addition to the CXR images. This is particularly important since CXR observations alone may not always be specific enough to reach a pneumonia or CHF clinical diagnosis. The normal class is determined by excluding any ICD-9 codes that may result in abnormalities visible on CXRs and also having no abnormal labels extracted from the relevant CXR reports using CXR report labeler \cite{wu2020cxrlabeler}. The code to run the inclusion and exclusion criteria is available on our \href{https://github.com/cxr-eye-gaze/eye-gaze-dataset}{GitHub repository}.

In addition, our sampling criteria prioritized the strategy for getting a good number of examples of disease features across a range of ages and sex from the source ED population. The goal is to support building and evaluation of computer vision algorithms that do not overly rely on age and sex biases, which may depict prominent visual features \cite{karargyris2019age}, to predict disease classes.

\subsubsection*{Preparation of Images}\zlabel{Preparation of images}
The 1,083 CXR images (\ztitleref{Inclusion/Exclusion} section) were converted from DICOM (Digital Imaging and Communications in Medicine) format to .png format: normalized (0-255), resized and padded to 1920x1080 to fit the radiologist's computer's monitor resolution (i.e., kept same aspect ratio) and to enable loading into Gazepoint Analysis UX Edition\cite{GazepointSoftware}.

\begin{figure}[!ht]
  \centering
 {\includegraphics[width=0.4\textwidth]{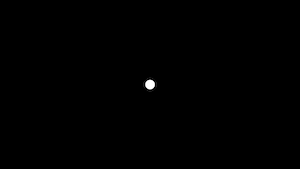}\label{fig:calibration}}
 \hspace{0.5cm}
 {\includegraphics[width=0.4\textwidth]{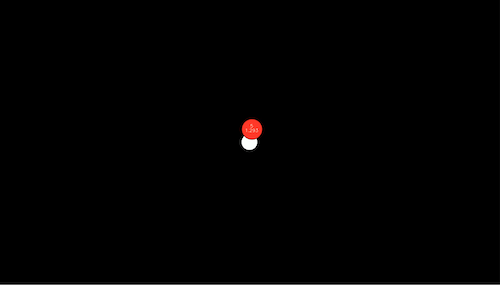}\label{fig:calibration-fixation}}
  \caption{Left: Calibration image presented to radiologist during data collection, Right: Radiologist's fixation over-imposed in red.}
  \label{fig:calibrations}

\end{figure}

A calibration image (i.e., resolution: 1920x1080 pixels) consisting of a white dot (30 pixels in radius) was generated (see Figure \ref{fig:calibrations} - left). The calibration image was presented to the radiologist randomly during data collection to measure eye gaze offset (see Figure \ref{fig:calibrations} - right).

The 1,083 images and calibration images were split into 38 numbered folders (i.e., '01', '02', '03',...'38') with no more than 30 images per folder. These folders were then uploaded to IBM's internal BOX\textsuperscript{TM} and shared with the radiologist who finally downloaded and loaded them to Gazepoint Analysis UX Edition software\cite{GazepointSoftware} to perform the reading (i.e., data collection).

\subsection*{Data Collection}

\subsubsection*{Software and Hardware Setup}\zlabel{Software Hardware Setup}
The Gazepoint GP3 Eye Tracker\cite{GazepointHardware} is an optical eye tracker that uses infrared light to detect eye fixation. The Gazepoint Analysis UX Edition software\cite{GazepointSoftware} is a software suite that comes along with the Gazepoint GP3 Eye Tracker and allows performing eye gaze experiments on image series. 

The Gazepoint GP3 Eye Tracker\cite{GazepointHardware} was set up in the radiologist's routine working environment on a Windows desktop PC computer (connected at USB 3 port). The Gazepoint Analysis UX Edition software\cite{GazepointSoftware} was also installed on the same computer. Each session was a standalone experiment that contained up to 30 images for reading by the radiologist. The radiologist's eyes were ~28 inches away from the monitor. The choice of this number of images was intentional to avoid fatigue and interruptions and to allow for timely offline review and quality assurance of each session recordings by the rest of the team.  Gazepoint Analysis UX Edition software\cite{GazepointSoftware} allows for 9-point calibration which occurred in the beginning of each session. In addition, Gazepoint Analysis UX Edition\cite{GazepointSoftware} allows the user to move to the next image either by pressing the spacebar on the keyboard when done with a case or by waiting for a fixed time. In this way the radiologist was able to move to the next CXR image when he was done with a given image, making the experiment easier.

\subsubsection*{Radiology Reading}
The radiologist read 1,083 CXR images reporting in unstructured prose, same as what he would perform in his routine working environment. The goal was to simulate a typical radiology read with minimal disruption from the eye gaze data collection process. The order of the CXR images was randomized to allow a blinded radiology read. In addition, we intentionally withheld the reason for exam information from our radiologist in order to collect an objective CXR exam interpretation based only on the available imaging features. 

The original source MIMIC-CXR Database \cite{johnson2019mimic} has the original de-identified free text reports for the same images, which were collected in real clinical scenarios where the reading radiologists had access to some patient clinical information outside the CXR image. The radiologists may even have had discussion about the patients with the bedside treating physician. Interpreting CXRs with additional patient clinical information (e.g., age, sex, other signs or symptoms) has the benefit of allowing radiologists to provide a narrower list of disease differential diagnosis by reasoning with their extra medical knowledge. However, it may also have the unintended effect of narrowing the radiology finding descriptions or subconsciously biasing what the radiologists look for in the image. In contrast, our radiologist only had the clinical information that all the CXRs came from an ED clinical setting.

By collecting a more objective read, we ensured that the CXR images used in this dataset have associated reports from both kinds of reading scenarios (read with and without patient clinical information). The goal is to broaden the range of possible technical and clinical research questions that future researchers working with the dataset may ask and explore.

\subsection*{Data Post-Processing}
\begin{figure}[!ht]
\centering
\includegraphics[scale=0.15]{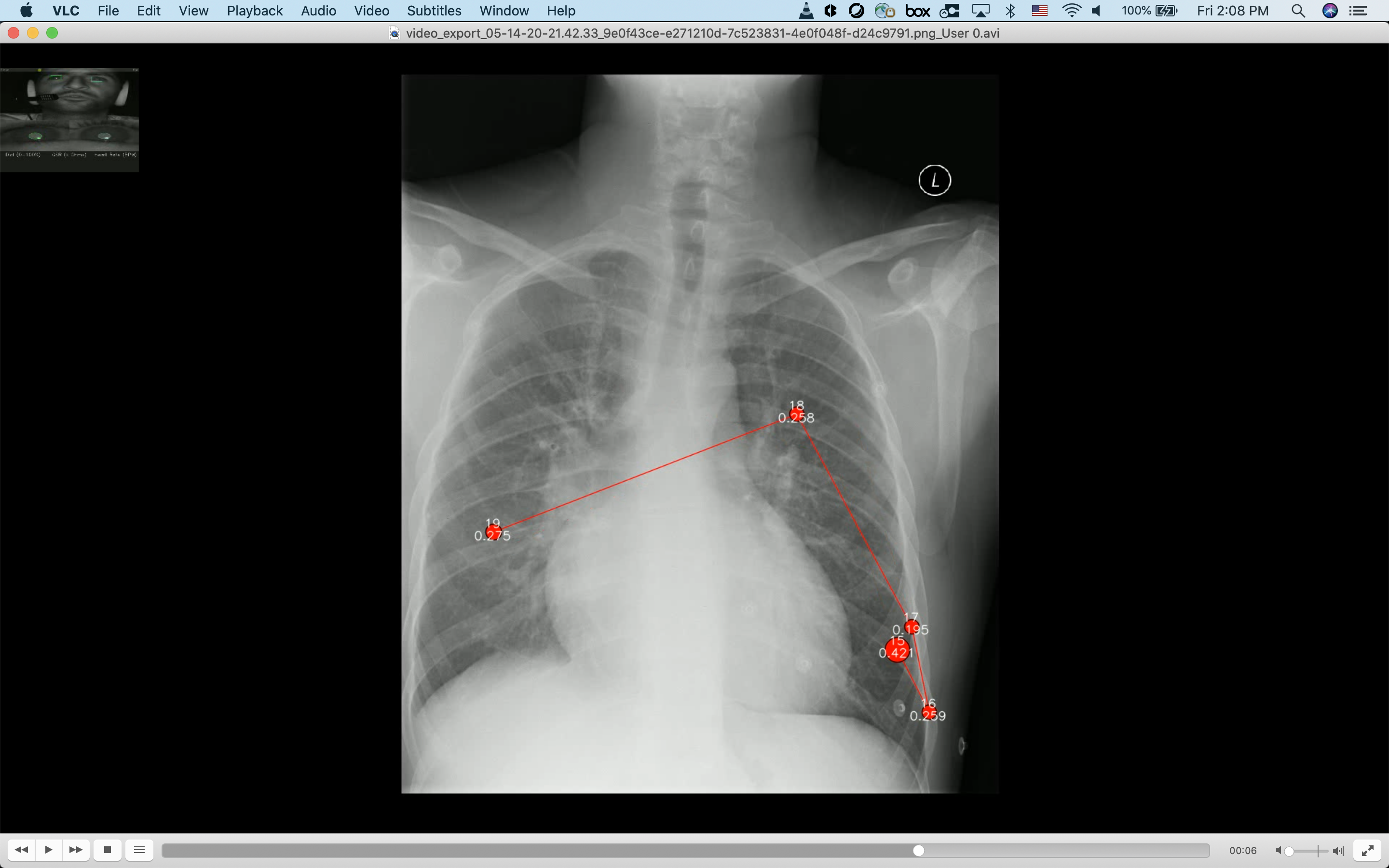}
\caption{Sample video exported from Gazepoint Analysis UX Edition\cite{GazepointSoftware} showing a CXR case image with overlayed fixations.}
\label{fig:video-sample}
\end{figure}

At the end of each session the radiologist exported the following information from the Gazepoint Analysis UX Edition software\cite{GazepointSoftware}: 1) fixation spreadsheet (.csv) containing fixation information for each case in the session, 2) eye gaze spreadsheet (.csv) containing raw eye gaze information for each case in the session, and 3) videos files (.avi) containing audio (i.e. radiologist's dictation) along with his eye gaze fixation heatmaps per session case (see Figure \ref{fig:video-sample}). These files were uploaded and shared over IBM's internal BOX\textsuperscript{TM} subscribed service. A team member reviewed each video for any technical quality issues (e.g., corrupted file, video playback stopped abruptly, bad audio quality).

\noindent Once data collection (i.e., 38 sessions) finished, the following post-processing tasks were performed.

\subsubsection*{Spreadsheet Merging}
From all sessions (i.e., folders), the fixations spreadsheets were concatenated into a single spreadsheet file: 
\texttt{fixations.csv}, and the raw eye gaze spreadsheets were concatenated into a single spreadsheet file: \texttt{eye\_gaze.csv}. 
Mapping of eye gaze and fixation from screen coordinate system to the original MIMIC image coordinate system was also performed at this stage. 

\noindent Detailed descriptions of these tables are provided in the \ztitleref{Data Records} section.

\subsubsection*{Audio Extraction and Transcript Generation}\zlabel{audio extraction}
For each session video file (i.e., containing radiologist's eye gaze fixations and dictation in .avi format, Figure \ref{fig:video-sample}) the dictation audio was extracted and saved in \texttt{audio.wav} and \texttt{audio.mp3} files. We used \href{https://cloud.google.com/speech-to-text}{Google Speech-to-Text} to transcribe the audio (i.e., wav file) into text. Transcribed text was saved in \texttt{transcript.json} containing timestamps and corresponding words based on the API example found in  \href{https://cloud.google.com/speech-to-text/docs/async-time-offsets}{documentation}. Furthermore, the transcripts were corrected manually by three (3) team members (all verified by the radiologist) using the original audio. An example of a transcript json is given in the \ztitleref{Data Records} section.

\subsubsection*{Segmentation Maps and Bounding Boxes for Anatomies}
\begin{figure}[!ht]
  \centering
 {\includegraphics[width=0.15\textwidth]{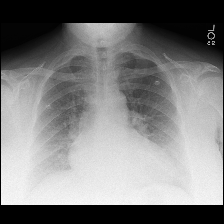}\label{fig:segmentation_map_original}}
 {\includegraphics[width=0.15\textwidth]{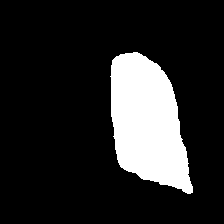}\label{fig:segmentation_map_left_lung}}
  {\includegraphics[width=0.15\textwidth]{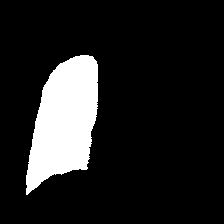}\label{fig:segmentation_map_right_lung}}
   {\includegraphics[width=0.15\textwidth]{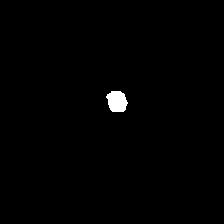}\label{fig:segmentation_map_aortic}}
    {\includegraphics[width=0.15\textwidth]{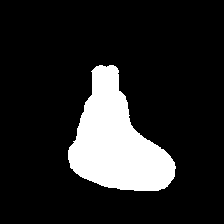}\label{fig:segmentation_map_mediastanum}}
    
  \caption{From Left to Right: CXR image, Left lung, Right lung, Aortic knob and Mediastinum.}
  \label{fig:segmentation_maps}
\end{figure}

\begin{figure}[!ht]
\centering
\includegraphics[scale=0.45]{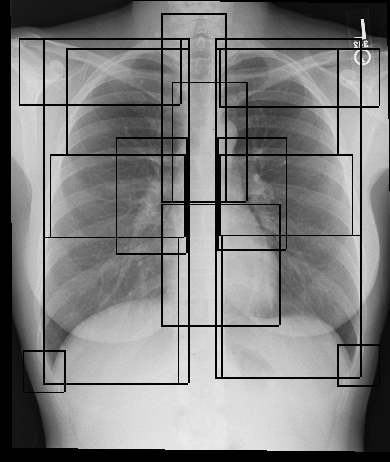}
\caption{Sample CXR case with 17 overlaying anatomical bounding boxes. The anatomies in the chest overlay one another on CXRs since the image is the 2D X-ray shadow capture of a 3D object.}
\label{fig:bbox-sample}
\end{figure}

Two supplemental datasets are also provided to enrich this dataset: 
\begin{itemize}
  \item \textbf{Segmentation maps:} \zlabel{Segmentation Maps} Four (4) key anatomical structures per image were generated: i) \texttt{left\_lung.png}, ii) \texttt{right\_lung.png}, iii) \texttt{mediastinum.png} and iv) \texttt{aortic\_knob.png}. These anatomical structures were automatically segmented by an internal segmentation model and then manually reviewed and corrected by the radiologist. Each image has pixel values 255 for anatomy and 0 for background. Figure \ref{fig:segmentation_maps} presents a sample case with its corresponding segmentation maps. 
  
  \item \textbf{Bounding boxes}: An extension of a bounding box extraction pipeline \cite{wu2020automatic} was used to extract 17 anatomical bounding boxes for each CXR image, which include: `right lung', `right upper lung zone', `right mid lung zone', `right lower lung zone', `left lung', `left upper lung zone', `left mid lung zone', `left lower lung zone', `right hilar structures', `left hilar structures', `upper mediastinum', `cardiac silhouette', `trachea', `right costophrenic angle', `left costophrenic angle', `right clavicle', `left clavicle'. These zones cover the clinically most important anatomies on a Posterior Anterior (PA) CXR image. These automatically produced bounding boxes were manually corrected (when required). Each bounding box is described by the top left corner point (X\textsubscript{X1}, Y\textsubscript{Y1})  and bottom right corner point (X\textsubscript{X2}, Y\textsubscript{Y2}) on the original CXR image coordinate system. Figure \ref{fig:bbox-sample} shows an example of anatomical bounding boxes. The information for bounding boxes of the 1,083 images are contained in \texttt{bounding\_boxes.csv}

\end{itemize}

Researchers can utilize these two (2) supplemental datasets to improve segmentation and disease localization algorithms by combining them with the eye gaze data. In the \ztitleref{statistical_analysis} subsection we utilize the \texttt{bounding\_boxes.csv} to perform statistical analysis between fixations and condition pairs.

\section*{Data Records}\zlabel{Data Records}

\begin{figure}[!ht]
\centering
\includegraphics[scale=0.5]{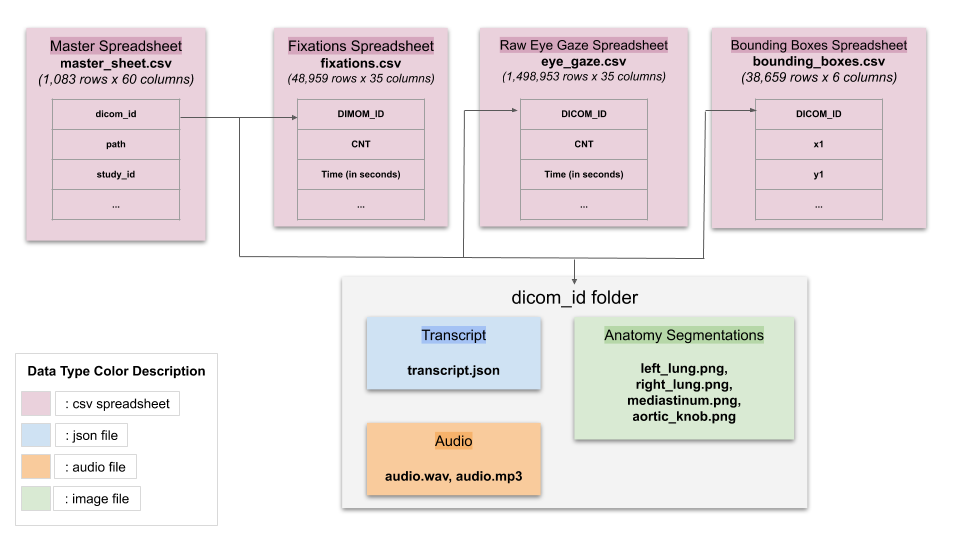}
\vspace{-0.5cm}
\caption{Overview of Dataset}
\label{fig:dataset-overview}
\end{figure}

An overview of the released dataset with their relationships is provided in Figure \ref{fig:dataset-overview}. Specifically four (4) data documents and one (1) folder are provided:

\begin{enumerate}
\item
\texttt{master\_sheet.csv}: Spreadsheet containing MIMIC DICOM ids along with study clinical indication sentence, report derived finding labels, and ICD-9 derived outcome disease labels.
\item

\texttt{eye\_gaze.csv}: Spreadsheet containing raw eye gaze data as exported by Gazepoint Analysis UX Edition software.\cite{GazepointSoftware}
\item

\texttt{fixations.csv}: Spreadsheet containing fixation data as exported by Gazepoint Analysis UX Edition software.\cite{GazepointSoftware}
\item

\texttt{bounding\_boxes.csv}: Spreadsheet containing bounding box coordinates for key frontal CXR anatomical structures.
\item

\texttt{audio\_segmentation\_transcripts}: Folder containing dictation audio files (i.e. mp3, wav), transcript file (i.e. json), anatomy segmentation mask files (i.e. png) for each dicom id.

\end{enumerate}

The dataset is hosted at \href{https://doi.org/10.13026/qfdz-zr67}{PhysioNet}\cite{eyegaze_dataset}. To utilize the dataset, the only requirement for the user is to obtain Physionet access to the MIMIC-CXR Database \cite{johnson2019mimic} in order to download the original MIMIC CXR images in DICOM format. The \texttt{dicom-id} tag found throughout all the dataset documents maps records to the MIMIC CXR images. 
A detailed description of each data document is provided in the following subsections. 

\subsubsection*{Master Spreadsheet}

The master spreadsheet (\texttt{master\_sheet.csv}) provides the following key information:
\begin{itemize}
\item 
The \texttt{dicom-id} column maps each row to the original MIMIC CXR image as well as the rest of the documents in this dataset. 
\item
The \texttt{study-id} column maps the CXR image/dicom to the associated CXR report, which can be found from the source MIMIC-CXR dataset\cite{johnson2019mimic}.
\item
For each CXR study (\texttt{study-id}), granular radiology `finding' labels have been extracted from the associated original MIMIC reports by two different NLP pipelines -- first is the CheXpert NLP pipeline\cite{irvin2019chexpert}, and second is an NLP pipeline developed internally\cite{wu2020cxrlabeler}.
\item
Additionally, for each CXR study (\texttt{study-id}), the reason for exam indication has been sectioned out from the original MIMIC CXR reports. The indication sentence(s) tend to contain patient clinical information that may not otherwise be visible from the CXR image alone.
\end{itemize}

Table \ref{long, tab:master_spreadsheet} describes in detail each column found in the master spreadsheet.

 \subsubsection*{Fixations and Eye Gaze Spreadsheets}\zlabel{fixation and eye gaze table}
The eye gaze information is stored in two (2) files: a) \texttt{fixations.csv}, and b) \texttt{eye\_gaze.csv}. Both files were exported by the Gazepoint Analysis UX Edition software\cite{GazepointSoftware}. 
Specifically, the \texttt{eye\_gaze.csv} file contains one row for every data sample collected from the eye tracker, while \texttt{fixations.csv} file contains a single data entry per fixation. The Gazepoint Analysis UX Edition software\cite{GazepointSoftware} generates the \texttt{fixations.csv} file from the \texttt{eye\_gaze.csv} file by averaging all data within a fixation to estimate the point of fixation based on the eye gaze samples, stopping when a saccade is detected.
Table \ref{long, tab:fixation_eye_gaze_spreadsheet} describes in detail each column found in the fixations and eye gaze spreadsheets.

 \subsubsection*{Bounding Boxes Spreadsheet}
The bounding boxes spreadsheet contains the following information:
\begin{itemize}
    \item 
    \texttt{dicom\_id}: DICOM ID as provided in MIMIC-CXR Database \cite{johnson2019mimic} for each image.
     \item 
    \texttt{bbox\_name}: These are the names for the 17 rectangular anatomical zones that bound the key anatomical organs on a frontal CXR image. Each lung (right and left) is bounded by its own bounding box, as well as subdivided into common radiological zones (upper, mid and lower lung zones) on each side. The upper mediastinum and the cardiac silhouette (heart) bounding boxes make up the mediastinum anatomy. The trachea is a bounding box that includes the visible tracheal air column on a frontal CXR, as well as the beginnings of the right and left main stem bronchi. The left and right hilar structures contain the left or right main stem bronchus as well as the lymph nodes and blood vessels that enter and leave the lungs in the hilar region. The left and right costophrenic angles are key regions to assess for abnormalities on a frontal CXR. The left and right clavicles can have potential fractures to rule out, but are also important landmarks to assess whether the patient (hence the anatomies on the CXRs) are rotated or not (which affects the appearance of potential abnormalities). Some of the bounding boxes (e.g clavicles) could be missing for an image if the target anatomical structure is cut off from the field of view of the CXR image.
        
     \item 
    \texttt{x1}: x coordinate for starting point of bounding box (upper left).
       \item 
    \texttt{y1}: y coordinate for starting point of bounding box (upper left).
       \item 
    \texttt{x2}: x coordinate for ending point of bounding box (lower right).
    \item 
    \texttt{y2}: y coordinate for ending point of bounding box (lower right).
    
\end{itemize}
 
Please see Figure \ref{fig:bbox-sample} for an example of all the anatomical bounding boxes. 

 \subsubsection* {Audio, Segmentation Maps and Transcripts}
 The \texttt{audio\_segmentation\_transcripts} folder contains subfolders for all the cases in the study with case \texttt{dicom\_id} as name. Each subfolder contains: a) the dictation audio file (mp3, wav), b) the segmentation maps of anatomies (png), as described in \ztitleref{Segmentation Maps} subsection above, and c) the dictation transcript (json). The dictation \texttt{transcript.json} contains the following tags: 
\begin{itemize}[leftmargin=*]
    \item 
    \texttt{full\_text}: The full text for the transcript.
    \item 
    \texttt{time\_stamped\_text}: The full text broken into timestamped phrases:
    \end{itemize}
\begin{itemize}

    \item
    \texttt{phrase}: Phrase text in the transcript.
    \indent \item 
    \texttt{begin\_time}: The starting time (in seconds) of dictation for a particular phrase.
    \item 
    \indent \texttt{end\_time}: The end time (in seconds) of dictation of a particular phrase.

\end{itemize}

\begin{figure}[!ht]
\centering
\includegraphics[scale=0.5]{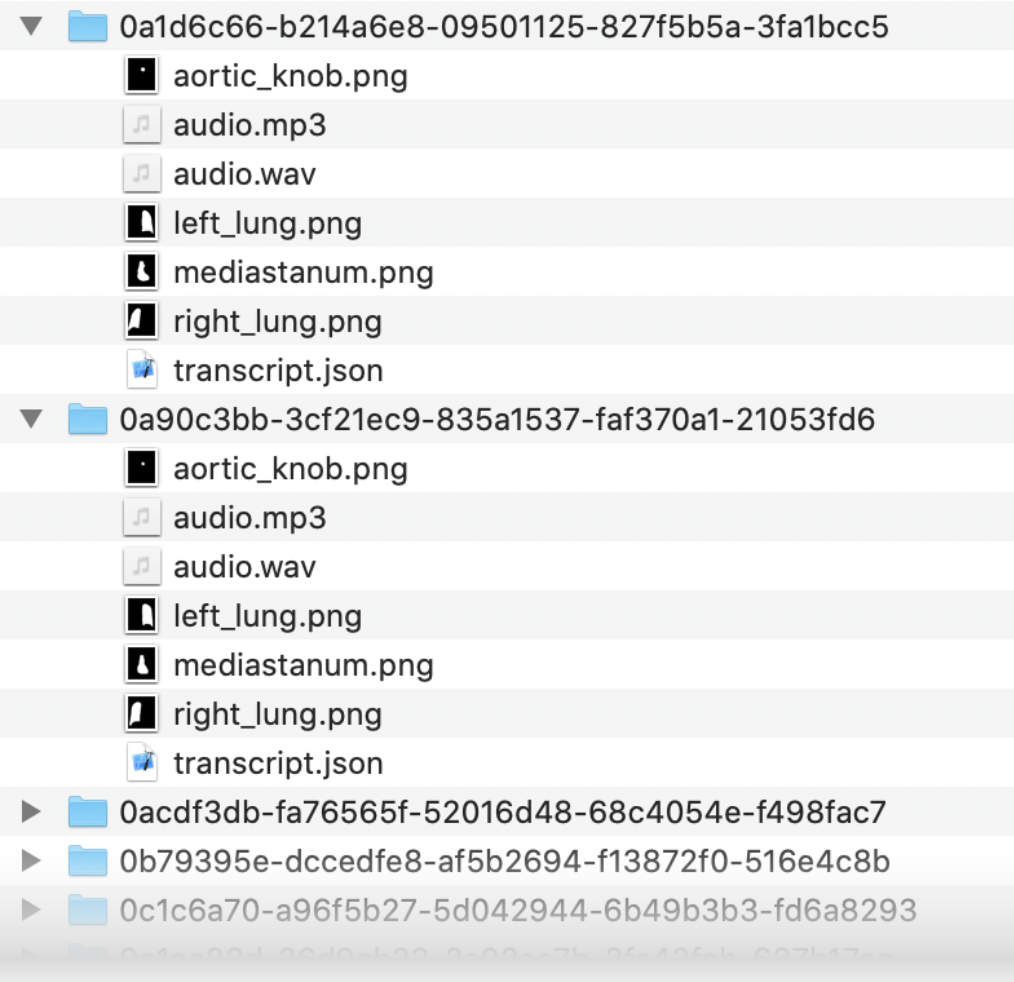}
\caption{\texttt{audio\_segmentation\_transcripts} folder structure}
\vspace{0.5cm}
\label{fig:folder_structure}
\end{figure}

Figure \ref{fig:folder_structure} shows the structure of the \texttt{audio\_segmentation\_transcripts} folder, while Figure \ref{fig:transcript_json} shows a transcript json example.

\section*{Technical Validation}
We subjected two aspects of the released data to reliability and quality validation: eye gaze and transcripts.

\noindent The code for the validation tasks below can be found on our \href{https://github.com/cxr-eye-gaze/eye-gaze-dataset}{GitHub repository}.

\subsubsection*{Validation of Eye Gaze Data}\zlabel{validation_eye_gaze}
As mentioned in the \ztitleref{Preparation of images} subsection, a calibration image was interjected randomly within the eye gaze sessions to measure the error of the eye gaze on the X- and Y- axis (Figure \ref{fig:calibrations}). 
A total of 59 calibration images were presented throughout the data collection. We calculated the error by using the fixation coordinates of the last entry of each calibration image (i.e. the final resting fixation by the radiologist on the calibration mark). The overall average percentage error on X, Y axes was calculated with (error\_X, error\_Y) = (0.0089 , 0.0504), and std: (0.0065, 0.0347) respectively. 
In pixels, the same error was: (error\_X, error\_Y) = (17.0356 , 54.3943), with std: (12.5529, 37.4257) respectively. 


\subsubsection*{Validation of Transcripts}
\begin{figure}[!ht]
\centering
\includegraphics[scale=0.4]{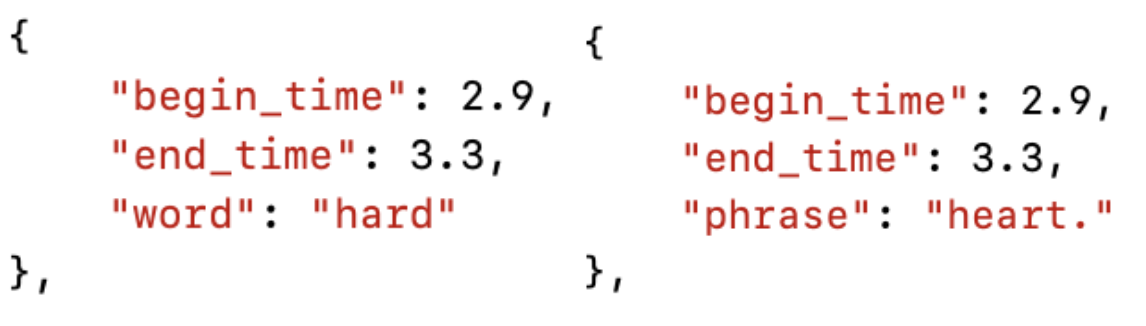}
\caption{Left: Example of incorrect detection. Right: Manual correction}
\label{fig:typeA_error}
\end{figure}

\begin{figure}[!ht]
\centering
\includegraphics[scale=0.6]{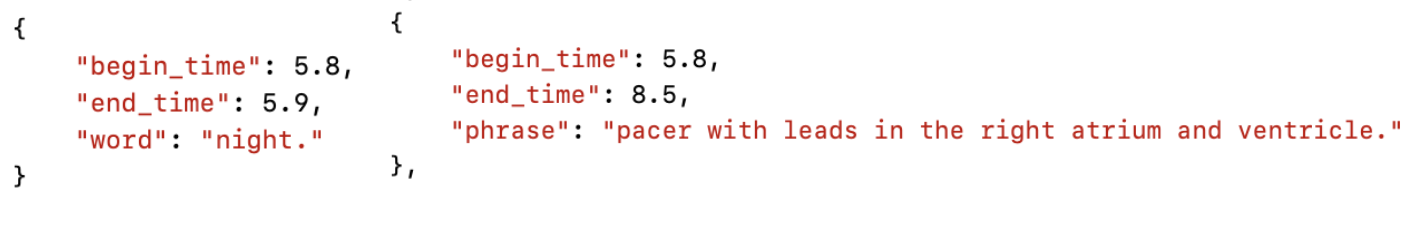}
\caption{Left: Missed and incorrect transcript phrase. Right: Manually corrected phrase}
\label{fig:typeB_error}
\end{figure}

\zlabel{validation_transcripts}
As mentioned in the \ztitleref{Preparation of images} subsection, transcripts were generated using \href{https://cloud.google.com/speech-to-text}{Google Speech-to-Text} on the dictation audio with timestamps per dictated word. The software produced two (2) types of errors: 
\begin{itemize}
    \item 
    Type A: Incorrect identification of a word at a particular time stamp (please see example in Figure \ref{fig:typeA_error}).
    \item
    Type B: Missed transcribed phrases of the dictation (please see example in Figure \ref{fig:typeB_error}).

\end{itemize}

The transcripts were manually corrected by three (3) experts and verified by the radiologist. Both types of errors were completely corrected. For Type B error, the missing text (i.e., more than one (1) word) was added with an estimation of the \texttt{begin\_time} and \texttt{end\_time} manually. To measure the potential error in the transcripts, the number of phrases with multiple words in a single time stamp was calculated (i.e., Type B error):

\begin{itemize}
    \item 
    Total number of phrases: 19,499
    \item
    Number of phrase with single words: 18,434
    \item
    Number of phrases with multiple words: 1,065
\end{itemize}
\[\text{Type B error}=1-\frac{1065}{19499}= 5.46\%\] 


\subsection*{Statistical Analysis on Fixations}\zlabel{statistical_analysis}
We performed t-test analysis to measure any significant differences between fixations for each condition within anatomical structures. More specifically, we performed the following steps:
\begin{enumerate}
    \item
    We examined the average number of fixations made in each disease condition, and found that the expert made significantly more overall fixations in the two diseased conditions than in the normal condition ($p<0.01$). 
    \item 
    For each image we calculated the number of fixations that their coordinates (i.e., X\_ORIGINAL, Y\_ORIGINAL in \texttt{fixations.csv}) fall into each anatomical zone (bounding box) found in \texttt{bounding\_boxes.csv}.  
    \item
    We performed t-test for each anatomical structure between condition pairs: i) Normal vs. Pneumonia, ii) Normal vs. CHF,  iii) Pneumonia vs CHF.
\end{enumerate}

\begin{figure}[!ht]
\centering
\includegraphics[scale=0.6]{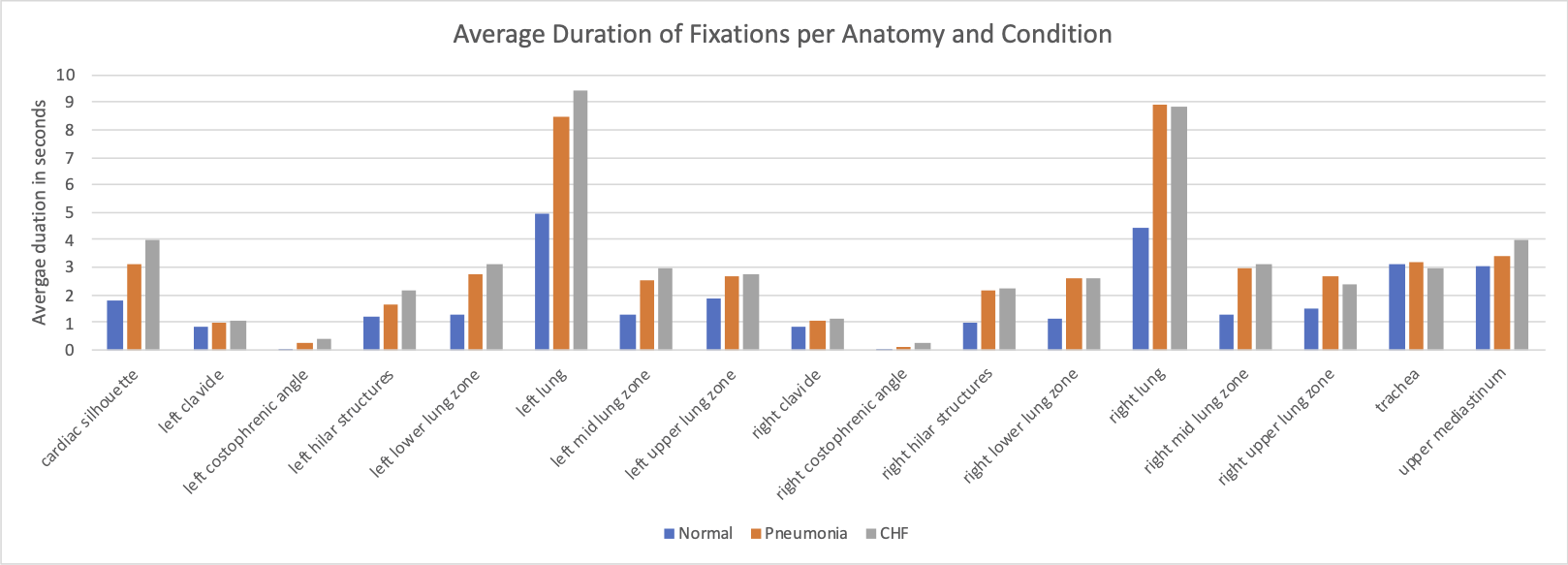}
\caption{Fixations vs. anatomical structures vs. conditions}
\label{fig:time_fixations_vs_anatomies}
\end{figure}

\begin{figure}[!ht]
\centering
\includegraphics[scale=0.5]{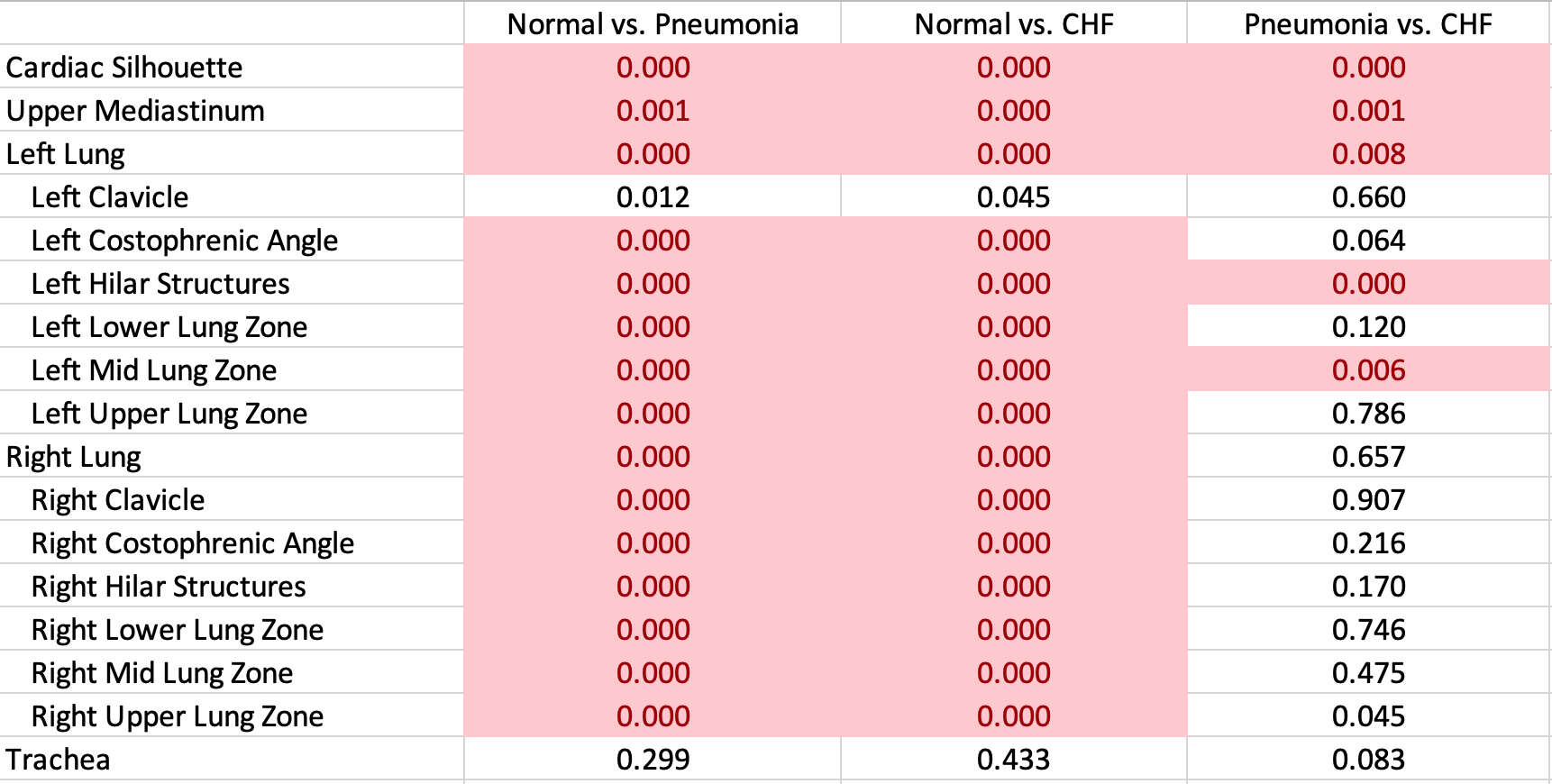}
\caption{p-values (at 3 decimal places) for each pair of condition and anatomy. p-values highlighted with red demonstrate statistical significant differences}
\label{fig:t-test}
\end{figure}

Figure \ref{fig:time_fixations_vs_anatomies} shows the duration of fixations per image for each disease condition and anatomical area, while Figure \ref{fig:t-test} shows p-values from each t-test. 
Fixations on Normal images are significantly different from Pneumonia and CHF. More fixations are made for images associated with either the Pneumonia of CHF final diagnoses. Moreover, fixations for the abnormal cases are mainly concentrated in anatomical regions (i.e., lungs and heart) that are relevant to the diagnosis, rather than distributed at random. Overall, the fixations on Pneumonia and CHF are comparatively similar, although still statistically different (e.g., Left Hilar Structure, Left Lung, Cardiac Silhouette, Upper Mediastinum). These statistical differences demonstrate that the radiologist's eye tracking information provides insight into the condition of the patient, and shows how a human expert pays attention to the relevant portions of the image when interpreting a CXR exam.
The code to replicate the t-test analysis can be found on the \href{https://github.com/cxr-eye-gaze/eye-gaze-dataset}{GitHub repository}.

\section*{Usage Notes}

The dataset is hosted on \href{https://doi.org/10.13026/qfdz-zr67}{PhysioNet}\cite{eyegaze_dataset}. The user is also required to apply for access to MIMIC-CXR Database \cite{johnson2019mimic} to download the images used in this study. 
Our \href{https://github.com/cxr-eye-gaze/eye-gaze-dataset}{GitHub repository} provides detailed description and source code (Python scripts) on how to use this dataset (e.g. post-processing, machine learning experiments) and reproduce the published validation results. 
The data in the MIMIC dataset has been previously de-identified, and the institutional review boards of the Massachusetts Institute of Technology (No. 0403000206) and Beth Israel Deaconess Medical Center (2001-P-001699/14) both approved the use of the database for research.


\subsection*{Use of the Dataset in Machine Learning}\zlabel{applications}
To demonstrate the effectiveness and richness of the information provided in this dataset, we performed two sets of machine learning multi-class classification experiments leveraging the eye-gaze data. These experiments are provided as dataset applications in simple and popular network architectures and they can function as a starting point for researchers. 

Both experiments used the eye gaze heatmap data to predict the multi-class classification of the aforementioned classes (i.e. Normal, CHF, Pneumonia in Table \ref{long, tab:master_spreadsheet}) and compare the performances with and without the eye gaze information. Our evaluation metric was AUC (Area Under the ROC Curve). The first experiment deal with leveraging information from the temporal eye gaze fixation heatmaps and the second uses static eye gaze fixation heatmaps. In contrast to temporal fixation heatmaps, static fixation heatmaps is the aggregation of all the temporal fixations into a single image.

\subsubsection*{Temporal Heatmaps Experiment}
\begin{figure}[!ht]
\centering
\includegraphics[scale=0.3]{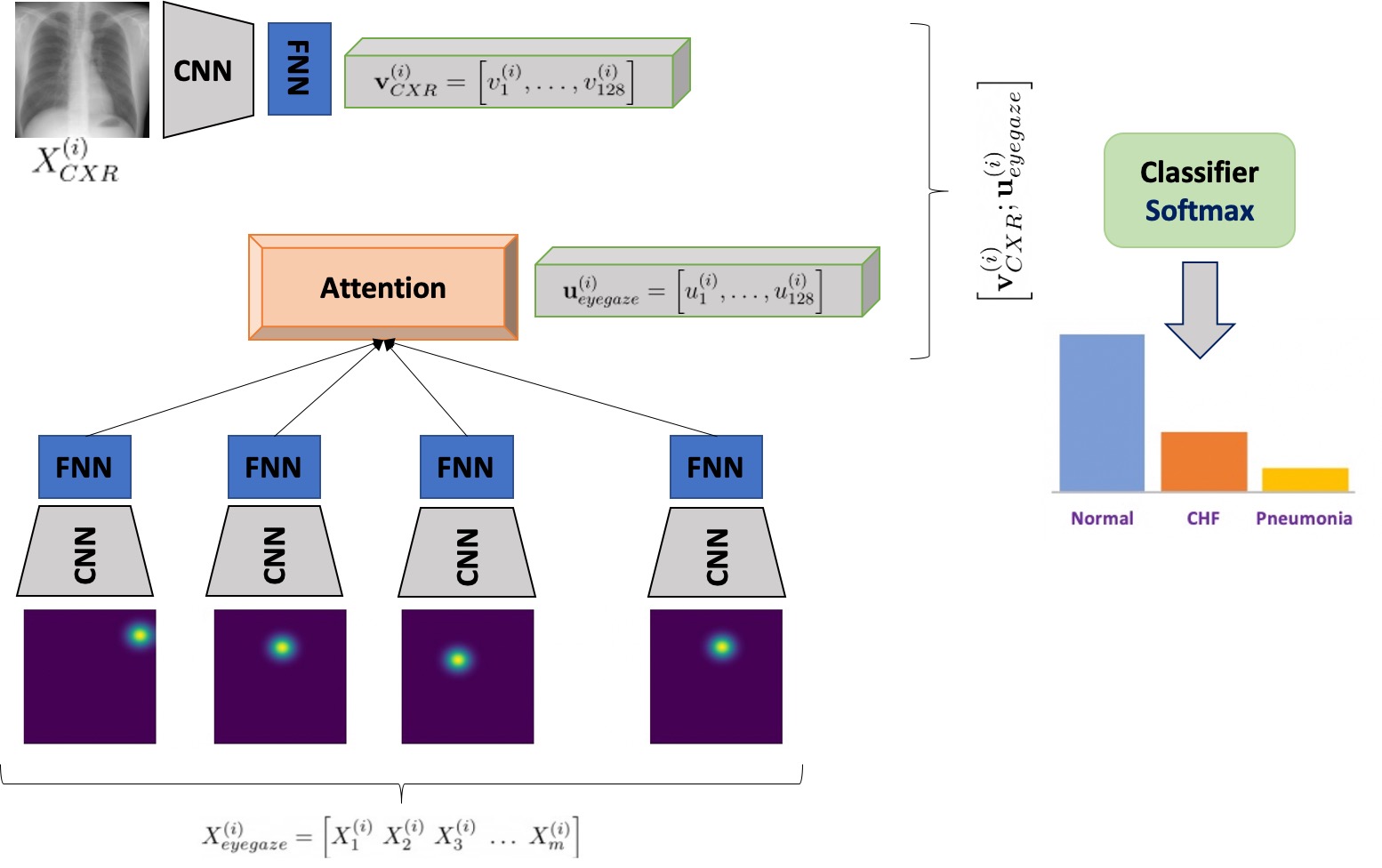}
\caption{Model architecture for leveraging temporal eyegaze information.}
\label{fig:model}
\end{figure}

\begin{figure}[!ht]
\centering
\includegraphics[scale=0.3]{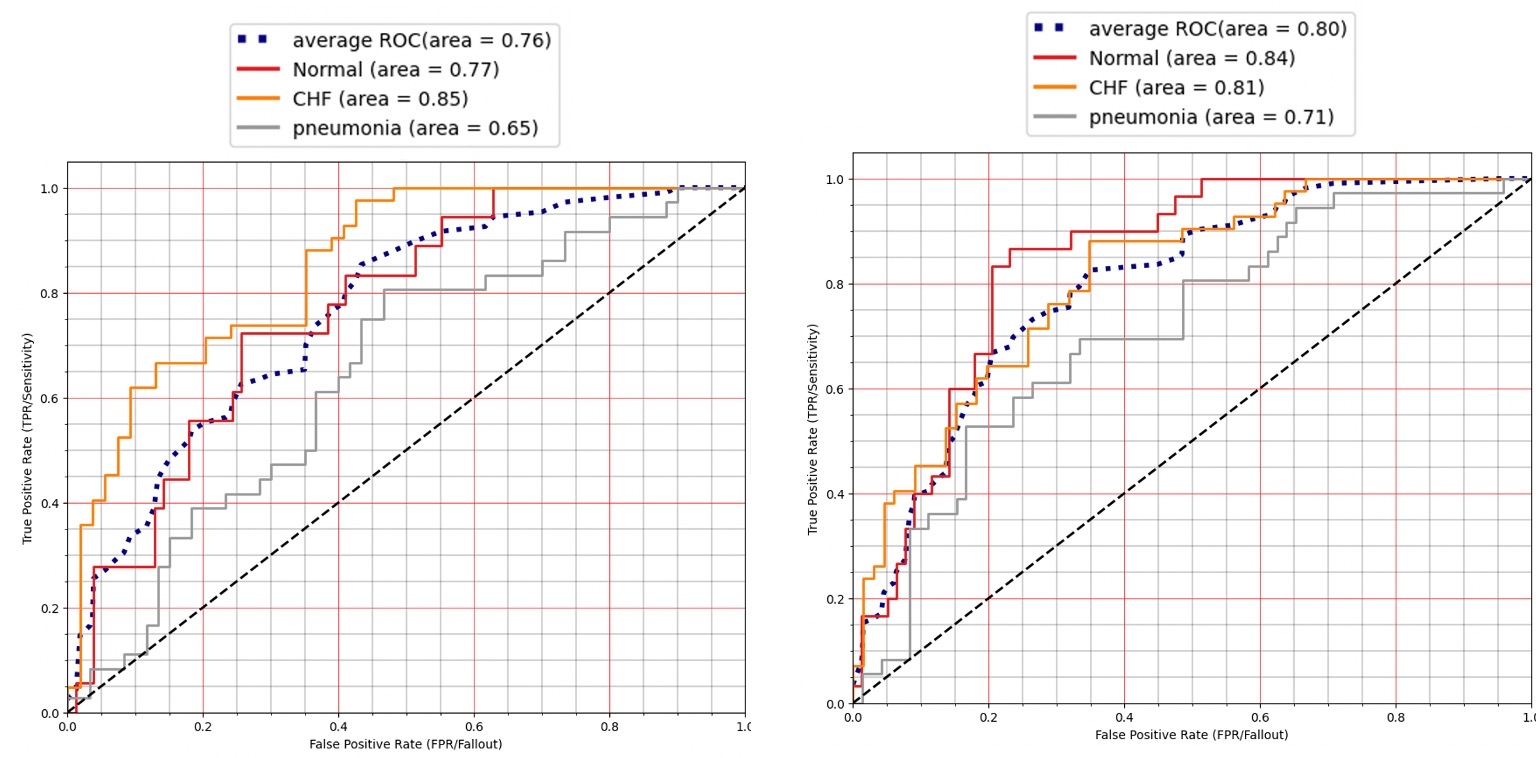}
\caption{Experimental results with (right) and without (left) temporal eyegaze information.}
\label{fig:exp1}
\end{figure}

The first model consists of a neural architecture, where the image and the temporal fixation heatmaps representations are concatenated before the final prediction layer. We denote an instance of this dataset as  $\mathit{X^{(i)}}$, which includes the image $X_{CRX}^{(i)}$ and the sequence of $m$ temporal fixation heatmaps $X_{eyegaze}^{(i)} = \{X_j^{(i)}\}_{k=1}^{m}$, where $k \in \{1, ..., m\}$ is the temporal heatmap index.
To acquire a fixed vector CRX representation $\mathbf{v}_{CRX}^{(i)}$, the image is passed through a convolutional layer with 64 filters of kernel size 7 and stride 2, followed by max-pooling, batch normalization and a dense layer of 128 units. The baseline model consists of the aforementioned image representation layer, combined with a final linear output layer that produces the classification prediction. Additionally, for the eye gaze, each heatmap is passed through a similar convolutional encoder and then the sequence of heatmaps is summarized with a 1-layer bidirectional LSTM with self-attention \cite{cheng2016long,vaswani2017attention}. We denote the heatmap representation as $\mathbf{u}_{eyegaze}^{(i)}$. Here, the image and heatmaps representations are concatenated before passed through the final classification layer. Figure \ref{fig:model} shows the full architecture. 
We train with Adam \cite{kingma2014adam}, $0.001$ initial learning rate and triangular schedule with fixed decay \cite{smith2017cyclical}, $16$ batch size and $0.5$ dropout \cite{srivastava2014dropout}. The experimental results in Figure \ref{fig:exp1} show that incorporating eye gaze temporal information, without any preprocessing, filtering or feature engineering, results in $4\%$ AUC improvement for this prediction task, when compared to the baseline model with just CXR image data as input.

\subsubsection*{Static Heatmaps Experiment}
\begin{figure}[!ht]
\centering
\includegraphics[scale=0.5]{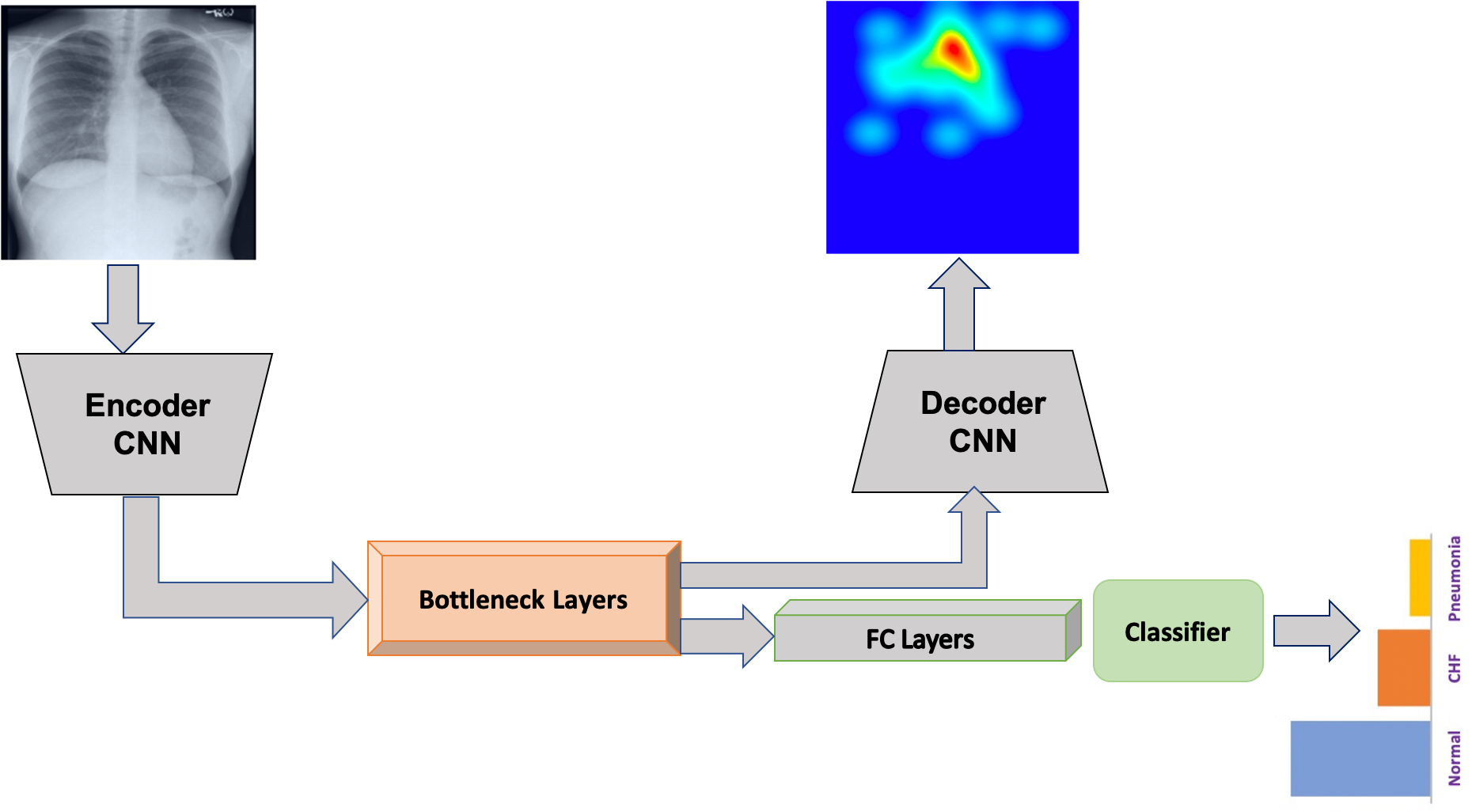}
\caption{Block diagram of U-Net utilizing the static heatmap combined with a classification head}
\label{fig:unet_block}
\end{figure}

The previous section showed the use of temporal fixation heatmaps with improvements demonstrated on a simple network architecture over baseline. In this experiment, we pose the classification problem in the U-Net architecture framework \cite{ronneberger2015u} with an additional multi-class classification block at the bottleneck layer (see Figure \ref{fig:unet_block}). The encoding and bottleneck arm of the U-Net can be any standard pre-trained classifier without the fully connected layer. The two combined will act as a feature encoder for the classifier and the CNN decoder part of the network runs deconvolution layers to predict the static eye gaze fixation heatmaps. The advantage is that we can jointly train to output the eye-gaze static fixation heatmap as well as predict the multi-class classification. Then, during testing on unseen CXR images, the network can predict the disease class and produce a probability heatmap of the most important locations pertaining to the condition.

 We used a pretrained EfficientNet-b0 \cite{tan2019efficientnet} as the encoder and bottleneck layers. The classification head was an adaptive average pooling followed by flatten, dropout \cite{srivastava2014dropout} and linear output layers. The decoder CNN consisted of three convolution followed by upsampling layers. The loss function was a weighted combination ($\gamma$) of the classification and the segmentation losses both of which used a binary cross entropy loss function. The baseline network consisted of just the encoder and the bottleneck arm followed by the classification head. 

\begin{figure}[!ht]
\centering
\includegraphics[scale=0.5]{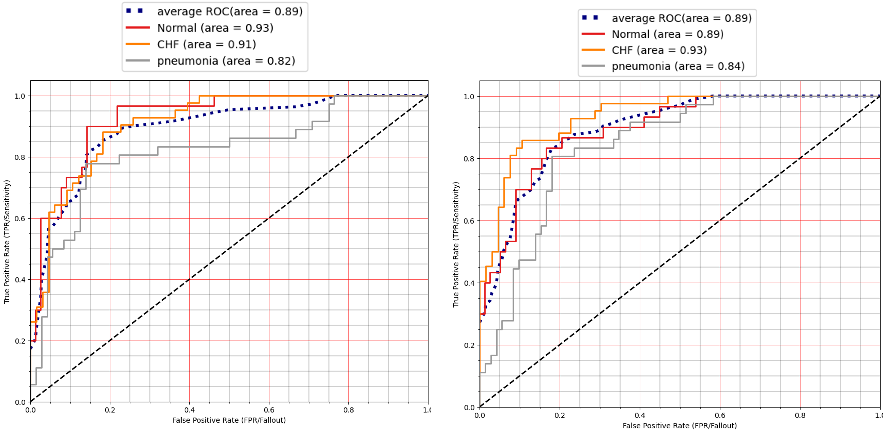}
\caption{AUC results with U-Net (right) and baseline classifier (left) static eyegaze information.}
\label{fig:exp2}
\end{figure}

\begin{figure}[!ht]
\begin{subfigure}{\textwidth}
  \centering
  \includegraphics[width=\textwidth]{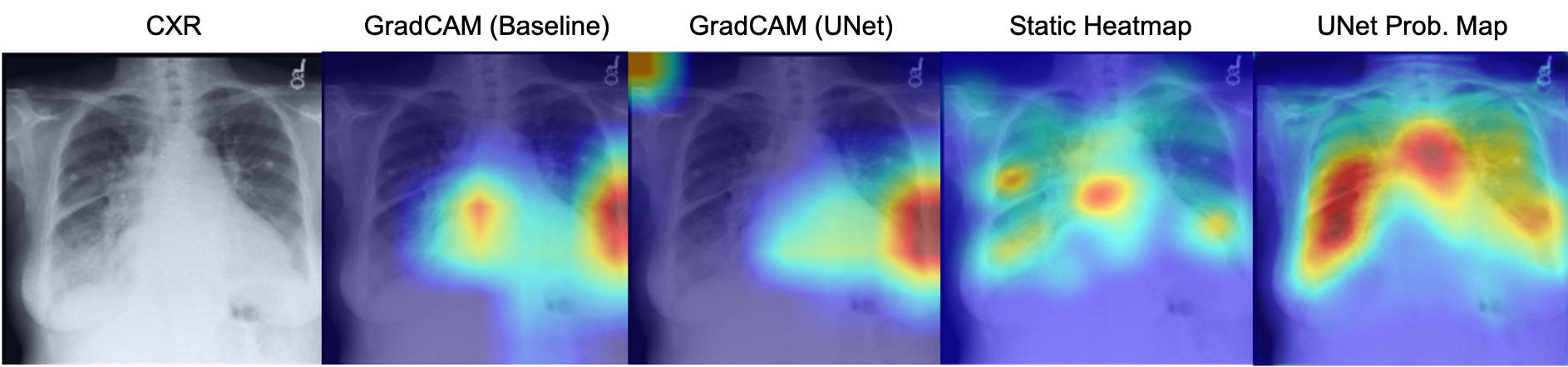}
  \caption{CHF. The physician's eye-gaze tends to fall on the enlarged heart and hila, as well as generally on the lungs}
  \label{fig:sfig1}
\end{subfigure}%

\begin{subfigure}{\textwidth}
  \centering
  \includegraphics[width=\textwidth]{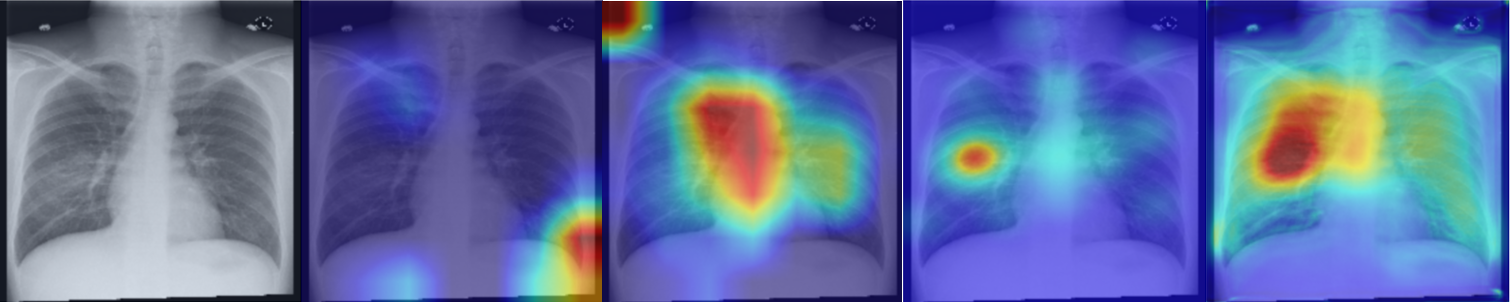}
  \caption{Pneumonia. The physician's eye-gaze predictably focuses on the focal lung opacity}
  \label{fig:sfig1}
\end{subfigure}%

\begin{subfigure}{\textwidth}
  \centering
  \includegraphics[width=\textwidth]{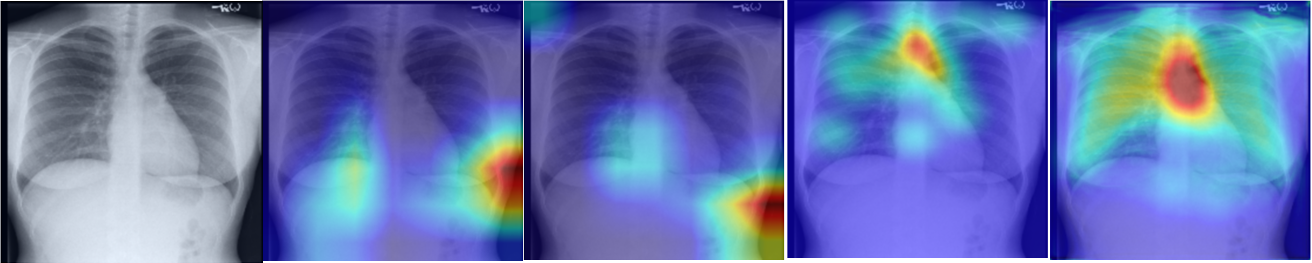}
  \caption{Normal. Because the lungs are clear, the physician's eye-gaze skips around the image without focus}
  \label{fig:sfig1}
\end{subfigure}%

  \caption{Qualitative comparison of the interpretabiltiy of U-Net based probability maps in comparison with GradCAM. From Left to Right: CXR image, GradCAM from Baseline Model, GradCAM from U-Net Encoder, Static EyeGaze Heatmap and U-Net Prob Map.}
  \label{fig:gradcam_images}
\end{figure}

The hyper-parameter tuning for both the U-Net and the baseline classifier was performed using the Tune library \cite{liaw2018tune} and the resulting best performing tune is shown in Table \ref{tab:tune}. Figure \ref{fig:exp2} shows the U-Net and baseline AUCs. Both had similar performance. However, for this experiment, we are interested in seeing how network interpretability improved with the use of static eye gaze heatmaps. Figure \ref{fig:gradcam_images} shows a qualitative comparison of the Grad-CAM \cite{selvaraju2017grad}. The Grad-CAM approach is one of the common methods to visualize activation maps of convolutional networks.  While the Grad-CAM based heatmaps don't clearly highlight the disease locations, we see clearly that the heatmap probability outputs of the U-Net highlights similar regions to what the static eye-gaze heatmap shows. 

With both experiments we tried to demonstrate different use cases of the eye gaze data into machine learning. With the first experiment we wanted to show how eye gaze data can be utilized in a human-machine setting where radiologist's eye gaze information are fed into the algorithm. The second experiment shows how eye eye gaze information can be used towards explainability purposes through generating verified activation maps. We intentionally did not include other modalities (audio, text) because of the complexity of such experiments and the scope of this paper (i.e., dataset description). We hope that these experiments can serve as a starting point for researchers to explore novel ways to utilize this multi-modal dataset.  

\section*{Limitations of study}
Although this study provides a unique large research dataset, we acknowledge the following limitations:
\begin{enumerate}
    \item 
   The study was performed with a single radiologist. This can certainly bias  the dataset (lacks inter-observer variability) and we aim to expand the data collection with multiple radiologists in the future. However, given the relatively large size and richness of data from various sources (i.e.  multi-modal), we believe that the current dataset already holds great value to the research community. In addition, we have shown with preliminary machine learning experiments that a model trained to optimize on a radiologist's eye tracking pattern has improved diagnostic performance as compared to a baseline model trained with weak image-level labels.
   
    \item
    The images used during the radiology reading were in 'png' format and not in DICOM. That's because the Gazepoint Analysis UX Edition\cite{GazepointSoftware} doesn't support DICOM format. This had the shortcoming that the radiologist could not utilize windowing techniques. However the png images were prepared using the windowing information in the original DICOM images.
    \item 
    This dataset includes only Posterior Anterior (PA) CXR images as selected from the inclusion/exclusion criteria (Figure \ref{fig:datasetflowchart}). This view position criterion was clinically chosen because of its higher quality images compared to Anterior Posterior (AP) CXR images. Therefore, any analysis (e.g., machine learning models trained on only this dataset) may suffer from generalizability to AP CXR images. 
   
\end{enumerate}

\section*{Code availability}

Our \href{https://github.com/cxr-eye-gaze/eye-gaze-dataset}{GitHub repository} contains code (Python 3) for:
\begin{enumerate}
    \item 
    Data Preparation
    \begin{itemize}
    \item 
    Inclusion and exclusion criteria on MIMIC dataset (see details in \ztitleref{Inclusion/Exclusion} section)
    \item
    Case sampling and image preparation for eye gaze experiment (see details in \ztitleref{Preparation of images} section)
    \end{itemize}
    \item
    Data Post -Processing
    \begin{itemize}
    \item 
    Speech-to-text on dictation audio (see details in \ztitleref{audio extraction} section)
    \item
    Mapping of eye gaze coordinates to original image coordinates (see details in \ztitleref{fixation and eye gaze table} section)
    \item
    Generate heatmap images (i.e temporal or static) and videos given eye gaze coordinates. The temporal and static heatmap images were used in our demonstrations of machine learning methods in \ztitleref{applications} section.
\end{itemize}
\item
Technical Validation
\begin{itemize}
    \item 
    Validation of eye gaze fixation quality using calibration images (see details in \ztitleref{validation_eye_gaze})
        \item 
    Validation of quality in transcribed dictations (see details in \ztitleref{validation_transcripts} section)
    \item 
    The t-test for eye gaze fixations for each anatomical structure and condition pairs (see details in \ztitleref{statistical_analysis} section)
    
        \end{itemize}
\item
Machine Learning Experiments, as described in \ztitleref{applications} section

\end{enumerate}

Software requirements are listed in the  \href{https://github.com/cxr-eye-gaze/eye-gaze-dataset}{GitHub repository}

\bibliography{bibliography}

\begin{thebibliography}{10}
\urlstyle{rm}
\expandafter\ifx\csname url\endcsname\relax
  \def\url#1{\texttt{#1}}\fi
\expandafter\ifx\csname urlprefix\endcsname\relax\def\urlprefix{URL }\fi
\expandafter\ifx\csname doiprefix\endcsname\relax\def\doiprefix{DOI: }\fi
\providecommand{\bibinfo}[2]{#2}
\providecommand{\eprint}[2][]{\url{#2}}

\bibitem{irvin2019chexpert}
\bibinfo{author}{Irvin, J.} \emph{et~al.}
\newblock \bibinfo{title}{{Chexpert: A large chest radiograph dataset with
  uncertainty labels and expert comparison}}.
\newblock In \emph{\bibinfo{booktitle}{{Proceedings of the AAAI Conference on
  Artificial Intelligence}, volume={33}, pages={590--597}, year={2019}}}.

\bibitem{johnson2019mimic}
\bibinfo{author}{Johnson, A.~E.} \emph{et~al.}
\newblock \bibinfo{journal}{\bibinfo{title}{{MIMIC-CXR, a de-identified
  publicly available database of chest radiographs with free-text reports}}}.
\newblock {\emph{\JournalTitle{Scientific Data}}} \textbf{\bibinfo{volume}{6}},
  \url{https://doi.org/10.13026/C2JT1Q} (\bibinfo{year}{2019}).

\bibitem{bluemke2020assessing}
\bibinfo{author}{Bluemke, D.~A.} \emph{et~al.}
\newblock \bibinfo{journal}{\bibinfo{title}{{{A}ssessing {R}adiology {R}esearch
  on {A}rtificial {I}ntelligence: {A} {B}rief {G}uide for {A}uthors,
  {R}eviewers, and {R}eaders-{F}rom the {R}adiology {E}ditorial {B}oard}}}.
\newblock {\emph{\JournalTitle{Radiology}}} \textbf{\bibinfo{volume}{294}},
  \bibinfo{pages}{487--489} (\bibinfo{year}{2020}).

\bibitem{waite2019analysis}
\bibinfo{author}{Waite, S.~A.} \emph{et~al.}
\newblock \bibinfo{journal}{\bibinfo{title}{{Analysis of perceptual expertise
  in radiology--Current knowledge and a new perspective}}}.
\newblock {\emph{\JournalTitle{Frontiers in human neuroscience}}}
  \textbf{\bibinfo{volume}{13}}, \bibinfo{pages}{213} (\bibinfo{year}{2019}).

\bibitem{van2017visual}
\bibinfo{author}{Van~der Gijp, A.} \emph{et~al.}
\newblock \bibinfo{journal}{\bibinfo{title}{{How visual search relates to
  visual diagnostic performance: a narrative systematic review of eye-tracking
  research in radiology}}}.
\newblock {\emph{\JournalTitle{Advances in Health Sciences Education}}}
  \textbf{\bibinfo{volume}{22}}, \bibinfo{pages}{765--787}
  (\bibinfo{year}{2017}).

\bibitem{krupinski2010current}
\bibinfo{author}{Krupinski, E.~A.}
\newblock \bibinfo{journal}{\bibinfo{title}{{Current perspectives in medical
  image perception}}}.
\newblock {\emph{\JournalTitle{Attention, Perception, \& Psychophysics}}}
  \textbf{\bibinfo{volume}{72}}, \bibinfo{pages}{1205--1217}
  (\bibinfo{year}{2010}).

\bibitem{tourassi2013investigating}
\bibinfo{author}{Tourassi, G.}, \bibinfo{author}{Voisin, S.},
  \bibinfo{author}{Paquit, V.} \& \bibinfo{author}{Krupinski, E.}
\newblock \bibinfo{journal}{\bibinfo{title}{{Investigating the link between
  radiologists' gaze, diagnostic decision, and image content}}}.
\newblock {\emph{\JournalTitle{Journal of the American Medical Informatics
  Association}}} \textbf{\bibinfo{volume}{20}}, \bibinfo{pages}{1067--1075}
  (\bibinfo{year}{2013}).

\bibitem{khosravan2019collaborative}
\bibinfo{author}{Khosravan, N.} \emph{et~al.}
\newblock \bibinfo{journal}{\bibinfo{title}{{A collaborative computer aided
  diagnosis (C-CAD) system with eye-tracking, sparse attentional model, and
  deep learning}}}.
\newblock {\emph{\JournalTitle{Medical image analysis}}}
  \textbf{\bibinfo{volume}{51}}, \bibinfo{pages}{101--115}
  (\bibinfo{year}{2019}).

\bibitem{stember2019eye}
\bibinfo{author}{Stember, J.~N.} \emph{et~al.}
\newblock \bibinfo{journal}{\bibinfo{title}{{Eye Tracking for Deep Learning
  Segmentation Using Convolutional Neural Networks}}}.
\newblock {\emph{\JournalTitle{Journal of digital imaging}}}
  \textbf{\bibinfo{volume}{32}}, \bibinfo{pages}{597--604}
  (\bibinfo{year}{2019}).

\bibitem{aresta2020automatic}
\bibinfo{author}{Aresta, G.} \emph{et~al.}
\newblock \bibinfo{journal}{\bibinfo{title}{{Automatic lung nodule detection
  combined with gaze information improves radiologists' screening
  performance}}}.
\newblock {\emph{\JournalTitle{IEEE Journal of Biomedical and Health
  Informatics}}}  (\bibinfo{year}{2020}).

\bibitem{mall2018modeling}
\bibinfo{author}{Mall, S.}, \bibinfo{author}{Brennan, P.~C.} \&
  \bibinfo{author}{Mello-Thoms, C.}
\newblock \bibinfo{journal}{\bibinfo{title}{{Modeling visual search behavior of
  breast radiologists using a deep convolution neural network}}}.
\newblock {\emph{\JournalTitle{Journal of Medical Imaging}}}
  \textbf{\bibinfo{volume}{5}}, \bibinfo{pages}{035502} (\bibinfo{year}{2018}).

\bibitem{goldberger2000physiobank}
\bibinfo{author}{Goldberger, A.~L.} \emph{et~al.}
\newblock \bibinfo{journal}{\bibinfo{title}{{PhysioBank, PhysioToolkit, and
  PhysioNet: components of a new research resource for complex physiologic
  signals}}}.
\newblock {\emph{\JournalTitle{circulation}}} \textbf{\bibinfo{volume}{101}},
  \bibinfo{pages}{e215--e220} (\bibinfo{year}{2000}).

\bibitem{johnson2020mimic4}
\bibinfo{author}{Johnson, A.} \emph{et~al.}
\newblock \bibinfo{title}{Mimic-iv}, \url{10.13026/A3WN-HQ05}
  (\bibinfo{year}{2020}).

\bibitem{GazepointHardware}
\bibinfo{author}{Gazepoint}.
\newblock \bibinfo{title}{{GP3 Eye Tracker}}.

\bibitem{GazepointSoftware}
\bibinfo{author}{Gazepoint}.
\newblock \bibinfo{title}{{Gazepoint Analysis UX Edition}}.

\bibitem{wu2020cxrlabeler}
\bibinfo{author}{Wu, J.~T.} \emph{et~al.}
\newblock \bibinfo{title}{{AI Accelerated Human-in-the-loop Structuring of
  Radiology Reports}}.
\newblock In \emph{\bibinfo{booktitle}{AMIA}} (\bibinfo{year}{2020}).

\bibitem{karargyris2019age}
\bibinfo{author}{Karargyris, A.} \emph{et~al.}
\newblock \bibinfo{title}{{Age prediction using a large chest x-ray dataset}}.
\newblock In \bibinfo{editor}{Mori, K.} \& \bibinfo{editor}{Hahn, H.~K.} (eds.)
  \emph{\bibinfo{booktitle}{Medical Imaging 2019: Computer-Aided Diagnosis}},
  vol. \bibinfo{volume}{10950}, \bibinfo{pages}{468 -- 476},
  \url{10.1117/12.2512922}. \bibinfo{organization}{International Society for
  Optics and Photonics} (\bibinfo{publisher}{SPIE}, \bibinfo{year}{2019}).

\bibitem{wu2020automatic}
\bibinfo{author}{{Wu}, J.} \emph{et~al.}
\newblock \bibinfo{title}{Automatic bounding box annotation of chest x-ray data
  for localization of abnormalities}.
\newblock In \emph{\bibinfo{booktitle}{2020 IEEE 17th International Symposium
  on Biomedical Imaging (ISBI)}}, \bibinfo{pages}{799--803}
  (\bibinfo{year}{2020}).

\bibitem{eyegaze_dataset}
\bibinfo{author}{Karargyris, A.} \emph{et~al.}
\newblock \bibinfo{title}{Eye gaze data for chest x-rays},
  \url{10.13026/QFDZ-ZR67} (\bibinfo{year}{2020}).

\bibitem{cheng2016long}
\bibinfo{author}{Cheng, J.}, \bibinfo{author}{Dong, L.} \&
  \bibinfo{author}{Lapata, M.}
\newblock \bibinfo{title}{{Long Short-Term Memory-Networks for Machine
  Reading}}.
\newblock In \emph{\bibinfo{booktitle}{Proceedings of the 2016 Conference on
  Empirical Methods in Natural Language Processing}}, \bibinfo{pages}{551--561}
  (\bibinfo{year}{2016}).

\bibitem{vaswani2017attention}
\bibinfo{author}{Vaswani, A.} \emph{et~al.}
\newblock \bibinfo{title}{{Attention is all you need}}.
\newblock In \emph{\bibinfo{booktitle}{Advances in neural information
  processing systems}}, \bibinfo{pages}{5998--6008} (\bibinfo{year}{2017}).

\bibitem{kingma2014adam}
\bibinfo{author}{Kingma, D.~P.} \& \bibinfo{author}{Ba, J.}
\newblock \bibinfo{title}{{Adam: A method for stochastic optimization}}.

\bibitem{smith2017cyclical}
\bibinfo{author}{Smith, L.~N.}
\newblock \bibinfo{title}{{Cyclical learning rates for training neural
  networks}}.
\newblock In \emph{\bibinfo{booktitle}{2017 IEEE Winter Conference on
  Applications of Computer Vision (WACV)}}, \bibinfo{pages}{464--472}
  (\bibinfo{organization}{IEEE}, \bibinfo{year}{2017}).

\bibitem{srivastava2014dropout}
\bibinfo{author}{Srivastava, N.}, \bibinfo{author}{Hinton, G.},
  \bibinfo{author}{Krizhevsky, A.}, \bibinfo{author}{Sutskever, I.} \&
  \bibinfo{author}{Salakhutdinov, R.}
\newblock \bibinfo{journal}{\bibinfo{title}{{Dropout: a simple way to prevent
  neural networks from overfitting}}}.
\newblock {\emph{\JournalTitle{The journal of machine learning research}}}
  \textbf{\bibinfo{volume}{15}}, \bibinfo{pages}{1929--1958}
  (\bibinfo{year}{2014}).

\bibitem{ronneberger2015u}
\bibinfo{author}{Ronneberger, O.}, \bibinfo{author}{Fischer, P.} \&
  \bibinfo{author}{Brox, T.}
\newblock \bibinfo{title}{U-net: Convolutional networks for biomedical image
  segmentation}.
\newblock In \emph{\bibinfo{booktitle}{International Conference on Medical
  image computing and computer-assisted intervention}},
  \bibinfo{pages}{234--241} (\bibinfo{organization}{Springer},
  \bibinfo{year}{2015}).

\bibitem{tan2019efficientnet}
\bibinfo{author}{Tan, M.} \& \bibinfo{author}{Le, Q.~V.}
\newblock \bibinfo{journal}{\bibinfo{title}{Efficientnet: Rethinking model
  scaling for convolutional neural networks}}.
\newblock {\emph{\JournalTitle{arXiv preprint arXiv:1905.11946}}}
  (\bibinfo{year}{2019}).

\bibitem{liaw2018tune}
\bibinfo{author}{Liaw, R.} \emph{et~al.}
\newblock \bibinfo{journal}{\bibinfo{title}{Tune: A research platform for
  distributed model selection and training}}.
\newblock {\emph{\JournalTitle{arXiv preprint arXiv:1807.05118}}}
  (\bibinfo{year}{2018}).

\bibitem{selvaraju2017grad}
\bibinfo{author}{Selvaraju, R.~R.} \emph{et~al.}
\newblock \bibinfo{title}{Grad-cam: Visual explanations from deep networks via
  gradient-based localization}.
\newblock In \emph{\bibinfo{booktitle}{Proceedings of the IEEE international
  conference on computer vision}}, \bibinfo{pages}{618--626}
  (\bibinfo{year}{2017}).

\end{thebibliography}

\section*{Author contributions statement}
A.K. was responsible for conceptualization, execution and management of this project.\\
S.K. contributed to the conceptualization and was responsible for the design and execution of the use case experiments, the data and code open sourcing and preparation of the manuscript. \\
I.L. designed the neural architecture that utilizes the temporal fixation heat maps, conducted the respective machine learning experiments, created the hyper-parameter tuning pipeline and edited the manuscript.\\
J.W. was responsible for the clinical design of the dataset, data sampling and documentation for the project, development of the bounding box pipeline and organizing the validation of results, as well as writing and editing the manuscript.\\
A.S. was responsible for interpretation of images with eye tracking, experimental design, and editing of the manuscript. \\
M.T. contributed prior experience with eye tracking studies, equipment, and methodologies, and assisted with the analysis.\\
S.A. contributed to the conceptualization and was responsible for data collection and processing\\
D.B. contributed to the original team formation and consulted on eye gaze tracking technology and data collection.\\
V.M. contributed to the initial concept review and securing funding for equipment/capital and execution of the project. \\
E.K. contributed to the clinical design of the study.\\
M.M. contributed to the design of experiments, supervision of the work and writing of the manuscript. \\

All authors reviewed the manuscript.

\section*{Competing interests}
The authors declare no conflicts of interest.


\section*{Figures \& Tables}

\begin{longtable}[c]{| L | L |}
\caption{Online-only Master Spreadsheet\label{long, tab:master_spreadsheet}}\\
 \hline
 \hline
 \multicolumn{1}{|m{4cm}}{Column Name} & \multicolumn{1}{|m{10cm}|}{Description}\\
 \hline
  \hline
 \endfirsthead
 \hline
 \endlastfoot
  \multicolumn{1}{|m{4cm}}{dicom-id} & \multicolumn{1}{|m{10cm}|}{DICOM ID in the original MIMIC dataset \cite{johnson2019mimic}} \\
   \hline
\multicolumn{1}{|m{4cm}}{path} & \multicolumn{1}{|m{10cm}|}{Path of DICOM image in the original MIMIC dataset} \\
 \hline
  \multicolumn{1}{|m{4cm}}{study-id} & \multicolumn{1}{|m{10cm}|}{Study ID in the original MIMIC dataset} \\
 \hline
  \multicolumn{1}{|m{4cm}}{patient-id} & \multicolumn{1}{|m{10cm}|}{Patient ID in the original MIMIC dataset} \\
 \hline
   \multicolumn{1}{|m{4cm}}{stay-id} & \multicolumn{1}{|m{10cm}|}{Stay ID in the original MIMIC dataset} \\
 \hline
  \multicolumn{1}{|m{4cm}}{gender} & \multicolumn{1}{|m{10cm}|}{Gender of patient in the original MIMIC dataset} \\
 \hline
   \multicolumn{1}{|m{4cm}}{anchor-age} & \multicolumn{1}{|m{10cm}|}{Age range in years of patient in the original MIMIC dataset} \\
 \hline
 \multicolumn{1}{|m{4cm}}{image-top-pad, image-bottom-pad, image-left-pad, image-right-pad} & \multicolumn{1}{|m{10cm}|}{Padding (top, bottom, left, right respectively) in pixels applied after re-scaling MIMIC image to 1920x1080} \\
  \hline

    \multicolumn{1}{|m{4cm}}{normal-reports} & \multicolumn{1}{|m{10cm}|}{No affirmed abnormal finding labels or descriptors documented in the original MIMIC-CXR reports, extracted using an internal CXR labeling pipeline\cite{wu2020cxrlabeler}.} \\
 \hline
 
    \multicolumn{1}{|m{4cm}}{Normal} & \multicolumn{1}{|m{10cm}|}{No abnormal chest related final diagnosis from the Emergency Department (ED) discharge ICD-9 records AND have normal-reports as defined above.} \\
 \hline
     \multicolumn{1}{|m{4cm}}{CHF} & \multicolumn{1}{|m{10cm}|}{A clinical diagnosis of heart failure (includes ICD-9 for congestive heart failure, chronic or acute on chronic heart failure) from the ED visit as determined from the associated ICD-9 discharge diagnostic code.} \\
 \hline
     \multicolumn{1}{|m{4cm}}{Pneumonia} & \multicolumn{1}{|m{10cm}|}{A clinical diagnosis of any lung infection (pneumonia) including bacterial and viral, as determined from the ICD-9 discharge diagnosis code of the ED visit.} \\
 \hline
        
  \multicolumn{1}{|m{4cm}}{dx1, dx2, dx3, dx4, dx5, dx6, dx7, dx8, dx9} & \multicolumn{1}{|m{10cm}|}{The descriptive ICD-9 diagnosis name associated with the Emergency Room admission for which the CXR study was ordered. ICD-9 final diagnoses are used to identify the 3 classes in the eye-gaze analysis and experiments.} \\
   \hline

 \multicolumn{1}{|m{4cm}}{dx1-icd, dx2-icd, dx3-icd, dx4-icd, dx5-icd, dx6-icd, dx7-icd, dx8-icd, dx9-icd} & \multicolumn{1}{|m{10cm}|}{ICD-9 code for corresponding dx } \\
  
   \hline

 \multicolumn{1}{|m{4cm}}{consolidation,	enlarged-cardiac-silhouette,linear-patchy-atelectasis,lobar-segmental-collapse,not-otherwise-specified-opacity,pleural-parenchymal-opacity,	pleural-effusion-or-thickening,	pulmonary-edema-hazy-opacity,	normal-anatomically, elevated-hemidiaphragm, hyperaeration,	vascular-redistribution} & \multicolumn{1}{|m{10cm}|}{Abnormal finding labels derived from the original MIMIC-CXR reports by an internal IBM CXR report labeler\cite{wu2020cxrlabeler}. 0: Negative, 1: Positive} \\

   \hline

    \multicolumn{1}{|m{4cm}}{atelectasis-chx, cardiomegaly-chx	consolidation-chx, edema-chx,enlarged-cardiomediastinum-chx	fracture-chx,lung-lesion-chx,lung-opacity-chx,no-finding-chx,pleural-effusion-chx,pleural-other-chx,pneumonia-chx,pneumothorax-chx,support-devices-chx} & \multicolumn{1}{|m{10cm}|}{ChexPert \cite{irvin2019chexpert} report derived abnormal finding labels for MIMIC-CXR. 0: negative, 1: positive, -1: uncertain} \\

  \hline
  
    \multicolumn{1}{|m{4cm}}{cxr\_exam\_indication} & \multicolumn{1}{|m{10cm}|}{The reason for exam sentences sectioned out from Indication section of the original MIMIC-CXR reports\cite{irvin2019chexpert}. They briefly summarizes patients immediate clinical symptoms, prior medical conditions and or recent procedures that are relevant for interpreting the CXR study within the clinical context.} \\
 
 \end{longtable}

\begin{longtable}[c]{| c | L |}
 \caption{Online-only Fixations and Eye Gaze Spreadsheets\label{long, tab:fixation_eye_gaze_spreadsheet}}\\

 \hline
 \hline
 Data Type/Column Name & \multicolumn{1}{m{10cm}|}{  Description}\\
 \hline
  \hline

 \endfirsthead

 \hline

 \endlastfoot

 DICOM-ID & \multicolumn{1}{m{10cm}|}{DICOM ID from original MIMIC dataset.} \\
 \hline
 CNT & \multicolumn{1}{m{10cm}|}{The counter data variable is incremented by 1 for each data record sent by the server. Useful to determine if any data packets are missed by the client.} \\
 \hline
 TIME(in secs) & \multicolumn{1}{m{10cm}|}{The time elapsed in seconds since the last system initialization or calibration. The time stamp is recorded at the end of the transmission of the image from camera to computer. Useful for synchronization and to determine if the server computer is processing the images at the full frame rate. For a 60 Hz camera, the TIME value should increment by 1/60 seconds.}\\
 \hline
 TIMETICK(f=10000000) & \multicolumn{1}{m{10cm}|}{This is a signed 64-bit integer which indicates the number of CPU time ticks for high precision synchronization with other data collected on the same CPU.} \\
 \hline
 FPOGX &   \multicolumn{1}{m{10cm}|}{The X- coordinates of the fixation POG, as a fraction of the screen size. (0,0) is top left, (0.5,0.5) is the screen center, and (1.0,1.0) is bottom right.}\\
 \hline
 FPOGY &  \multicolumn{1}{m{10cm}|}{The Y-coordinates of the fixation POG, as a fraction of the screen size. (0,0) is top left, (0.5,0.5) is the screen center, and (1.0,1.0) is bottom right.}\\
 \hline
 FPOGS & \multicolumn{1}{m{10cm}|}{The starting time of the fixation POG in seconds since the system initialization or calibration.} \\
 \hline
 FPOGD & \multicolumn{1}{m{10cm}|}{The duration of the fixation POG in seconds}\\
 \hline
 FPOGID & \multicolumn{1}{m{10cm}|}{The fixation POG ID number}\\
 \hline 
 FPOGV & \multicolumn{1}{m{10cm}|}{The valid flag with value of 1 (TRUE) if the fixation POG data is valid, and 0 (FALSE) if it is not. FPOGV valid is TRUE ONLY when either one, or both, of the eyes are detected AND a fixation is detected. FPOGV is FALSE all other times, for example when the subject blinks, when there is no face in the field of view, when the eyes move to the next fixation (i.e. a saccade)}\\
 \hline
 BPOGX & \multicolumn{1}{m{10cm}|}{The X-coordinates of the best eye POG, as a fraction of the screen size.}\\
 \hline
 BPOGY & \multicolumn{1}{m{10cm}|}{The Y-coordinates of the best eye POG, as a fraction of the screen size.} \\
 \hline
 BPOGV & \multicolumn{1}{m{10cm}|}{The valid flag with value of 1 if the data is valid, and 0 if it is not.} \\
 \hline
 LPCX & \multicolumn{1}{m{10cm}|}{The X-coordinates of the left eye pupil in the camera image, as a fraction of the camera image size.} \\
 \hline
 LPCY & \multicolumn{1}{m{10cm}|}{The Y-coordinates of the left eye pupil in the camera image, as a fraction of the camera image size.}\\
 \hline
 LPD & \multicolumn{1}{m{10cm}|}{The diameter of the left eye pupil in pixels}\\
 \hline
 LPS & \multicolumn{1}{m{10cm}|}{The scale factor of the left eye pupil (unitless). Value equals 1 at calibration depth, is less than 1 when user is closer to the eye tracker and greater than 1 when user is further away.}\\
 \hline
 LPV & \multicolumn{1}{m{10cm}|}{The valid flag with value of 1 if the data is valid, and 0 if it is not.}\\
 \hline
 RPCX & \multicolumn{1}{m{10cm}|}{The X-coordinates of the right eye pupil in the camera image, as a fraction of the camera image size.} \\
 \hline
 RPCY & \multicolumn{1}{m{10cm}|}{The Y-coordinates of the right eye pupil in the camera image, as a fraction of the camera image size.} \\
 \hline
 RPD & \multicolumn{1}{m{10cm}|}{The diameter of the right eye pupil in pixels}\\
 \hline
 RPS & \multicolumn{1}{m{10cm}|}{The scale factor of the right eye pupil (unitless). Value equals 1 at calibration depth, is less than 1 when user is closer to the eye tracker and greater than 1 when user is further away.}\\
 \hline
 RPV & \multicolumn{1}{m{10cm}|}{The valid flag with value of 1 if the data is valid, and 0 if it is not.}\\
 \hline
 BKID & \multicolumn{1}{m{10cm}|}{Each blink is assigned an ID value and incremented by one. The BKID value equals 0 for every record where no blink has been detected.}\\
 \hline
 BKDUR & \multicolumn{1}{m{10cm}|}{The duration of the preceding blink in seconds.} \\
 \hline
 BKPMIN & \multicolumn{1}{m{10cm}|}{The number of blinks in the previous 60 second period of time.}\\
 \hline
 LPMM & \multicolumn{1}{m{10cm}|}{The diameter of the left eye pupil in millimeters.}\\
 \hline
 LPMMV & \multicolumn{1}{m{10cm}|}{The valid flag with value of 1 if the data is valid, and 0 if it is not.}\\
 \hline
 RPMM & \multicolumn{1}{m{10cm}|}{The diameter of the right eye pupil in millimeters.}\\
 \hline
 RPMMV &  \multicolumn{1}{m{10cm}|}{The valid flag with value of 1 if the data is valid, and 0 if it is not.}\\
 \hline
 SACCADE-MAG & \multicolumn{1}{m{10cm}|}{Magnitude of the saccade calculated as distance between each fixation (in pixels).}\\
 \hline
 SACCADE-DIR & \multicolumn{1}{m{10cm}|}{The direction or angle between each fixation (in degrees from horizontal).}\\ 
 \hline
 X\_ORIGINAL & \multicolumn{1}{m{10cm}|}{The X coordinate of the fixation in original DICOM image.}\\ 
 \hline
 Y\_ORIGINAL & \multicolumn{1}{m{10cm}|}{The Y coordinate of the fixation in original DICOM image.}\\ 
 \end{longtable}

\begin{figure}[!ht]
\centering
 \begin{mdframed}[linecolor=black, topline=false, bottomline=false,
  leftline=false, rightline=false, backgroundcolor=white ,
  linecolor=black,%
  leftmargin =-1cm,
  rightmargin=+1cm,
  usetwoside=false,
]

\begin{minted}{json}
{
    "full_text": " prominent heart. cardiomegaly in fact. suspect small left effusion with atelectasis. right lung appears clear.",
    "time_stamped_text": [
        {
            "begin_tim": 1.0,
            "end_time": 1.7000000000000002,
            "phrase": "prominent"
        },
        {
            "begin_tim": 1.7000000000000002,
            "end_time": 2.0,
            "phrase": "heart."
        },
        {
            "begin_tim": 3.0,
            "end_time": 3.8,
            "phrase": "cardiomegaly"
        },
        {
            "begin_tim": 3.8,
            "end_time": 4.2,
            "phrase": "in fact."
        },
        {
            "begin_tim": 6.3,
            "end_time": 7.1,
            "phrase": "suspect"
        },
        {
            "begin_tim": 8.2,
            "end_time": 8.9,
            "phrase": "small"
        },
        {
            "begin_tim": 8.9,
            "end_time": 9.0,
            "phrase": "left"
        },
        {
            "begin_tim": 9.0,
            "end_time": 9.4,
            "phrase": "effusion"
        },
        {
            "begin_tim": 10.7,
            "end_time": 11.2,
            "phrase": "with"
        },
        {
            "begin_tim": 11.2,
            "end_time": 11.6,
            "phrase": "atelectasis."
        },
        {
            "begin_tim": 17.9,
            "end_time": 18.3,
            "phrase": "right"
        },
        {
            "begin_tim": 18.3,
            "end_time": 18.5,
            "phrase": "lung"
        },
        {
            "begin_tim": 18.5,
            "end_time": 18.6,
            "phrase": "appears"
        },
        {
            "begin_tim": 18.6,
            "end_time": 19.0,
            "phrase": "clear."
        }
    ]
}
\end{minted}
\end{mdframed}
\caption{Example of transcript}
\label{fig:transcript_json}
\end{figure}

\begin{table}[h]
\centering
\caption{Best performing hyper-parameters used for the static heatmap experiments found using the Tune \cite{liaw2018tune} library.}
\label{tab:tune}
\resizebox{\textwidth}{!}{%
\begin{tabular}{@{}|l|l|l|l|l|l|l|@{}}
\toprule
\multicolumn{1}{|c|}{\textbf{\begin{tabular}[c]{@{}c@{}}Experiment\\  Name\end{tabular}}} & \textbf{Optimizer} & \multicolumn{1}{c|}{\textbf{\begin{tabular}[c]{@{}c@{}}Initial \\ Learning rate\end{tabular}}} & \textbf{\begin{tabular}[c]{@{}l@{}}Scheduler \cite{smith2017cyclical}\\ Step Size\end{tabular}} & \textbf{Epochs} & \textbf{Dropout\cite{srivastava2014dropout}} & \textbf{$\gamma$} \\ \midrule
UNet & Adam \cite{kingma2014adam} & 0.0091 & 2 & 35 & 0.5 & 0.417 \\ \midrule
Baseline Classifier & Adam \cite{kingma2014adam} & 0.0065 & 8 & 20 & 0.0 & N/A \\ \bottomrule
\end{tabular}%
}
\end{table}



\end{document}